\RequirePackage{fix-cm}
\documentclass{article}
\usepackage{iclr2027_conference,times}

\usepackage{amsmath,amsfonts,bm}

\def\eqref#1{equation~\ref{#1}}

\def\1{\bm{1}}

\DeclareMathAlphabet{\mathsfit}{\encodingdefault}{\sfdefault}{m}{sl}
\SetMathAlphabet{\mathsfit}{bold}{\encodingdefault}{\sfdefault}{bx}{n}

\usepackage{amsmath}
\usepackage{booktabs}
\usepackage{array}
\usepackage{colortbl}
\usepackage{graphicx}
\usepackage{float}
\usepackage{placeins}
\usepackage{algorithm}
\usepackage{algorithmic}
\usepackage{multirow}
\usepackage{xcolor}
\usepackage{hyperref}
\usepackage{silence}
\usepackage{microtype}
\usepackage{url}
\usepackage{listings}
\usepackage[skins,breakable,listings]{tcolorbox}

\definecolor{promptaccent}{HTML}{157A6E}
\definecolor{promptbg}{HTML}{F2F8F7}
\definecolor{caseaccent}{HTML}{B85C16}
\definecolor{casebg}{HTML}{FFF6EC}
\lstdefinestyle{mcgpoprompt}{
  basicstyle=\rmfamily\normalsize,
  breaklines=true,
  breakatwhitespace=true,
  columns=fullflexible,
  keepspaces=true,
  showstringspaces=false,
  aboveskip=0pt,
  belowskip=0pt
}
\tcbset{mcgpobox/.style={
  enhanced,
  breakable,
  coltitle=white,
  fonttitle=\rmfamily\bfseries\normalsize,
  boxrule=0.8pt,
  arc=1.5mm,
  outer arc=1.5mm,
  boxsep=0.5mm,
  left=1.5mm,
  right=1.5mm,
  top=1.2mm,
  bottom=1.2mm,
  before skip=0.55\baselineskip,
  after skip=0.55\baselineskip
}}
\newtcblisting{promptbox}[1]{
  mcgpobox,
  listing only,
  listing engine=listings,
  listing options={style=mcgpoprompt},
  title={#1},
  title after break={#1 (continued)},
  colback=promptbg,
  colframe=promptaccent,
  colbacktitle=promptaccent
}
\newtcblisting{casebox}[1]{
  mcgpobox,
  listing only,
  listing engine=listings,
  listing options={style=mcgpoprompt},
  title={#1},
  title after break={#1 (continued)},
  colback=casebg,
  colframe=caseaccent,
  colbacktitle=caseaccent
}
\newtcolorbox{rolloutbox}[1]{
  mcgpobox,
  title={#1},
  title after break={#1 (continued)},
  fontupper=\rmfamily\normalsize,
  colback=casebg,
  colframe=caseaccent,
  colbacktitle=caseaccent,
  before upper={\setlength{\parskip}{2pt}\setlength{\parindent}{0pt}}
}

\ifdefined\pdfsuppressptexinfo
\fi
\ifdefined\pdfobjcompresslevel
\fi

\makeatletter
\ifdefined\pdfrunninglinkoff
  \protected\def\pdfannot@link@off@@{\pdfrunninglinkoff}
  \protected\def\pdfannot@link@on@@{\pdfrunninglinkon}
  \ExplSyntaxOn
  \cs_new_eq:NN \mcgpo_shipout_picture_overlay:n \__shipout_picture_overlay:n
  \cs_set:Npn \__shipout_picture_overlay:n #1
    {
      \pdfrunninglinkoff
      \mcgpo_shipout_picture_overlay:n {#1}
      \pdfrunninglinkon
    }
  \ExplSyntaxOff
\fi
\makeatother

\title{Matched-Context Group Policy Optimization\\for LLM Agent Training}

\author{%
  \mbox{Chenxu Zhang\textsuperscript{1,3,*}},
  \mbox{Yaorui Shi\textsuperscript{2,3}},
  \mbox{Wenlong Zhang\textsuperscript{2}},
  \mbox{Jianchao Tan\textsuperscript{2}},
  \mbox{Zihan Wang\textsuperscript{3,4}, Yiming Cheng\textsuperscript{4}}%
}

\newcommand{\mcgpoaffiliations}{%
  \textsuperscript{1}Imperial College London,\space
  \textsuperscript{2}University of Science and Technology of China\\[1pt]
  \textsuperscript{3}Meituan,\space
  \textsuperscript{4}Tsinghua University%
}

\newcommand*{\mcgpocorrespondingemail}{cz3625@ic.ac.uk}
\newcommand{\mcgpoauthornote}{%
  \textsuperscript{*}Corresponding author.%
  \ifx\mcgpocorrespondingemail\empty\else
    \ \href{mailto:\mcgpocorrespondingemail}{\texttt{\mcgpocorrespondingemail}}%
  \fi
}

\hypersetup{
  pdftitle={Matched-Context Group Policy Optimization for LLM Agent Training},
  pdfauthor={Chenxu Zhang, Yaorui Shi, Wenlong Zhang, Jianchao Tan, Zihan Wang, Yiming Cheng},
  pdfsubject={Matched-context preference construction for LLM agent training},
  pdfkeywords={MCGPO, reinforcement learning, language model agents, credit assignment}
}

\newcommand{\method}{MCGPO}
\newcommand{\graphgpo}{\textsc{GraphGPO}}
\definecolor{tableaccent}{HTML}{4F79B8}
\definecolor{tablegroup}{HTML}{EEF3FA}
\definecolor{tableours}{HTML}{F2F2F2}
\definecolor{tablebest}{HTML}{F5D5D9}
\definecolor{tablesecond}{HTML}{DCE5FA}

\definecolor{RCBlue}{HTML}{5A78B4}
\definecolor{RCOrange}{HTML}{E17C05}
\definecolor{RCGain}{HTML}{055C29}
\definecolor{RCLoss}{HTML}{9B1C1C}
\definecolor{RCTablePanel}{HTML}{CFDFEE}
\definecolor{RCTableModel}{HTML}{EEF3FA}
\definecolor{RCTableGray}{HTML}{E5E9ED}
\definecolor{RCTableGrayText}{HTML}{4B5563}
\definecolor{RCTableBlue}{HTML}{DCE8F7}
\definecolor{RCTableBlueText}{HTML}{3F5F8F}
\definecolor{RCTableOrange}{HTML}{FCE5C5}
\definecolor{RCTableOrangeText}{HTML}{A85200}
\definecolor{RCObsBg}{HTML}{F4F7FA}
\definecolor{RCObsBorder}{HTML}{7891AA}
\definecolor{RCMethodRow}{HTML}{F2F2F2}
\definecolor{RCBaselineRow}{HTML}{FFFFFF}

\usepackage{mcgpo_preprint}

\begin{document}
\maketitle
\newcommand{\qwenFourBase}{78.18}
\newcommand{\qwenFourOurs}{82.14}
\newcommand{\qwenFourGain}{3.96}
\newcommand{\qwenFourShaping}{79.85}
\newcommand{\qwenFourShapingGain}{2.29}
\newcommand{\qwenEightBase}{79.69}
\newcommand{\qwenEightOurs}{84.84}
\newcommand{\qwenEightGain}{5.16}
\newcommand{\qwenEightShaping}{82.19}
\newcommand{\qwenEightShapingGain}{2.66}
\newcommand{\repAlfSmallStateGain}{8.08}
\newcommand{\repAlfSmallRubricGain}{5.84}
\newcommand{\repAlfLargeStateGain}{3.96}
\newcommand{\repAlfLargeRubricGain}{5.15}
\newcommand{\repWebSmallStateGain}{5.63}
\newcommand{\repWebSmallRubricGain}{7.71}
\newcommand{\repWebLargeStateGain}{9.90}
\newcommand{\repWebLargeRubricGain}{9.01}
\newcommand{\repStateMin}{3.96}
\newcommand{\repStateMax}{9.90}
\newcommand{\repRubricMin}{5.15}
\newcommand{\repRubricMax}{9.01}

\newcommand{\factorFoldWeightedGain}{2.34}
\newcommand{\factorWeightOppositeGain}{1.61}
\newcommand{\factorFoldUniformGain}{2.40}
\newcommand{\factorWeightSameGain}{1.67}

\begin{abstract}
Training long-horizon language model agents requires assigning credit to
intermediate decisions from sparse terminal feedback. Within a finite group of sibling rollouts,
useful action comparisons depend on their context, evidence for both
alternatives, and how that evidence enters learning. We propose
\emph{Matched-Context Group Policy Optimization (\method{})}, a procedure
for constructing weighted action preferences from already collected
trajectories. \method{} defines comparison contexts with pre-action state
summaries and observed rubric progress, annotated by task rules.
Two-action support selects among ordered exact and coarse matching rules.
Fixed folds separate candidate decisions from the terminal outcomes used
to orient their preferences, and support-dependent evidence weights
scale an auxiliary pairwise objective alongside shared GraphGPO task
learning. \method{} reuses rollout data and each response's
original history, preserving critic-free training. Across six
environment--backbone settings with three training seeds each, \method{}
improves mean success over matched GraphGPO: by 0.47--1.46
percentage points on ALFWorld and WebShop with Qwen2.5-1.5B and 7B, and
by \qwenFourGain{} and \qwenEightGain{} points on WebShop with Qwen3-4B and 8B. The ALFWorld
gains build on baseline success above 92\%, where remaining headroom is
limited. Integrated success improves in three of four Qwen2.5 settings.
Three-seed factorial controls support opposite-fold evidence and evidence
weighting; joint state--rubric matching achieves the highest mean in all
four representation-control settings.
\end{abstract}
\begingroup
\setlength{\intextsep}{4pt}
\begin{figure}[H]
\centering
\includegraphics[page=1,width=0.8\linewidth]{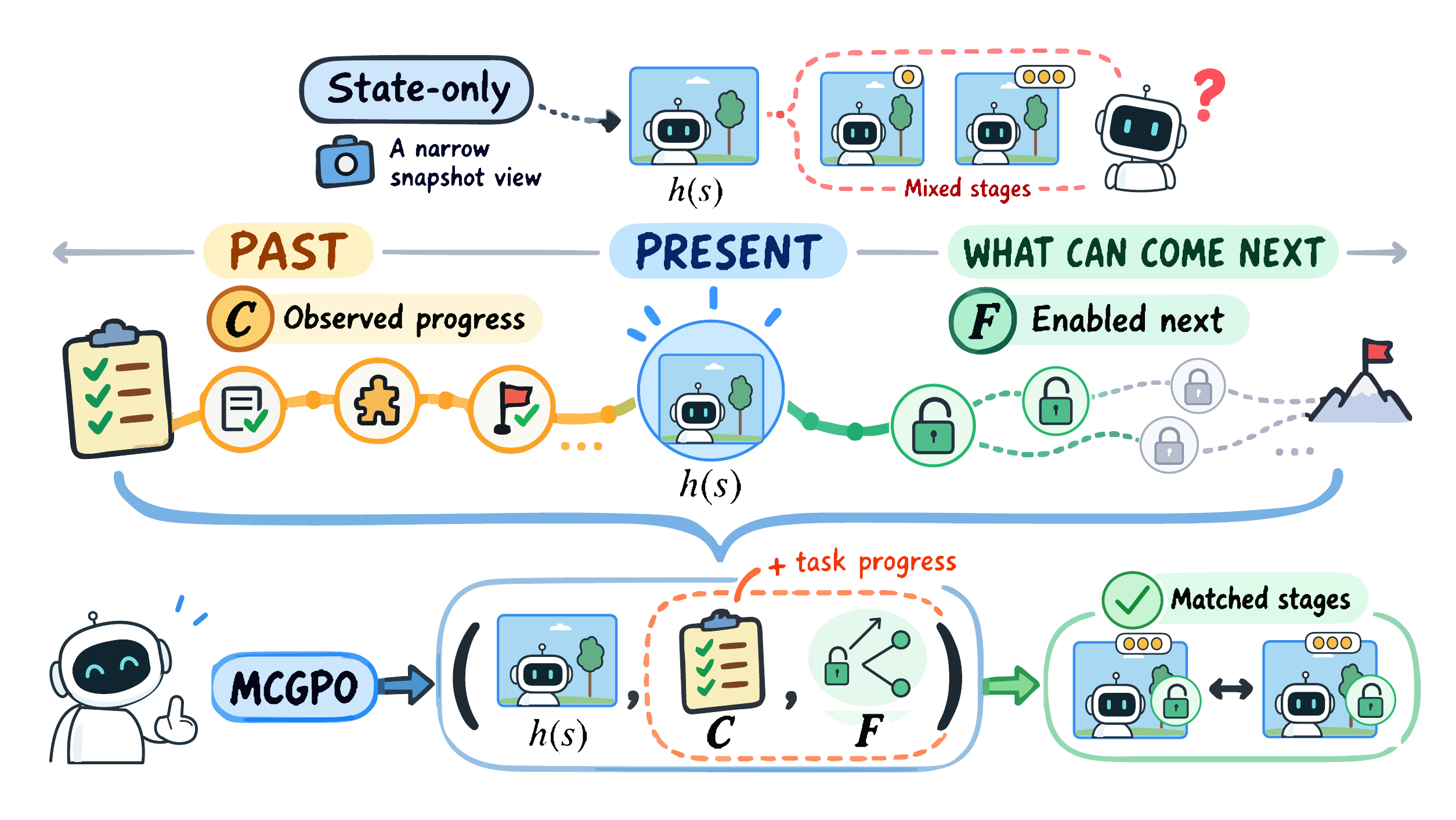}
\caption{Task context for local action comparisons. The same state summary
$h(s)$ can occur at different stages of a task. \method{} groups decisions
using the state summary and observed rubric progress $C$, with enabled
criteria annotated by the rule-derived frontier $F$.}
\label{fig:teaser}
\end{figure}
\endgroup

\section{Introduction}
\label{sec:introduction}

Training large language model agents requires \emph{credit assignment}:
sparse terminal rewards must guide the intermediate actions that shape
task outcomes \citep{zhou2024archer,chen2025loop,wang2025ragen}.
In ALFWorld \citep{shridhar2021alfworld} and WebShop
\citep{yao2022webshop}, success depends on a sequence of interdependent
decisions.
Group-relative reinforcement learning offers critic-free updates by
comparing outcomes within a group of sampled responses or trajectories
\citep{shao2024deepseekmath,ahmadian2024back,yu2025dapo}.
For multi-turn agents, GiGPO localizes these comparisons at recurring
anchor states \citep{feng2025gigpo}, while trajectory-graph methods
extract step credit from collected transitions
\citep{li2025salt,cheng2026graphgpo}.
However, grouping by repeated states or transitions can compare decisions
with different prior task progress, obscuring what their outcome contrast
represents.

This mismatch arises when \emph{the same state summary describes
different remaining decision problems}. A page, location, or inventory
summary may omit progress established before the current choice.
Consequently, an action taken after more prerequisites have been met can
inherit a higher success rate, while a useful action taken earlier can
receive weaker relative feedback. Their outcome contrast can mix prior
progress with the contribution of the current action. A captured
ALFWorld task illustrates this representational issue: two empty-handed
visits to the same countertop share a coarse state key, although one
follows acquiring and relocating a target object
(Appendix~\ref{app:resolution-census}). Adding context can separate such
decisions, but stricter conditions also reduce the repeated evidence
available for comparing different actions.

Recent work enriches comparison context to address this tension.
HGPO combines groups at multiple history depths,
and BiPACE uses policy representations and action-conditioned peer
baselines \citep{he2026hgpo,wang2026bipace}.
Task structure offers another source of context: explicit criteria guide reward
aggregation \citep{gunjal2025rar,lv2026gear}, policy stages and memory
\citep{li2026rubricem}, and milestone-level credit
\citep{wang2026beacon,qian2026milegpo}.
However, an explicit preference between different actions also needs
evidence for both alternatives. A richer context can leave one action
poorly supported, so a larger outcome gap may reflect only a few
trajectories. The remaining challenge is to select comparisons and
weight their influence using the evidence available within a finite
rollout group. This leads to our research question:

\begin{center}
\begingroup
\setlength{\fboxsep}{9pt}
\colorbox{black!4}{%
    \parbox{0.88\linewidth}{%
        \centering
        {\small\bfseries\scshape Research Question}\par
        \vspace{3pt}
        \emph{How can a finite group of existing rollouts supply supported local action preferences for policy learning?}%
    }%
}
\endgroup
\end{center}

We introduce \method{} (Matched-Context Group Policy Optimization) to
construct auxiliary action preferences from collected sibling rollouts.
The core idea is to select a comparison context supported by both actions
before forming a weighted preference. \textbf{Match} groups pre-action decisions by
state summaries and observed rubric progress, with a rule-derived
frontier annotating progress. It tries ordered exact and coarse state
rules, selecting the first with evidence for both actions.
\textbf{Label} uses terminal outcomes from the opposite fixed
fold to orient pairs and assign support-discounted evidence weights.
\textbf{Learn} optimizes a weighted pairwise objective alongside the
shared GraphGPO task objective. The labels express observational outcome
contrasts under the rollout policy; their weights control each pair's
optimization contribution. \method{} reuses rollout data and preserves
critic-free training and each response's original history.

We evaluate \method{} on ALFWorld and WebShop with Qwen2.5-1.5B and 7B,
using three training runs per setting. Mean success
improves over matched GraphGPO in all four settings by 0.47--1.46
percentage points, including ALFWorld baselines already above 92\%
success. A three-seed WebShop extension yields gains of \qwenFourGain{} and \qwenEightGain{}
points with Qwen3-4B and 8B, respectively. Integrated success over the
full training horizon improves in three Qwen2.5 settings, with positive
paired differences in each of their three seeds. Three-seed factorial controls
support opposite-fold evidence and evidence weighting. Across all four representation-control
settings, joint state--rubric conditioning raises three-seed mean success
over both State-only and Rubric-only matching. These results demonstrate
performance gains across the tested backbones and support joint
state--rubric conditioning as an effective comparison design.

Our contributions are threefold:
\begin{itemize}
    \item \textbf{Supported local preference construction.} We combine
    explicit comparison contexts, two-action support checks, and
    opposite-fold terminal evidence within an existing rollout group.
    \item \textbf{Evidence-weighted policy learning.} We connect these
    observational preferences to an auxiliary pairwise objective while
    retaining the shared task objective and original response histories.
    \item \textbf{Evaluation of the construction and its scope.} Matched
    comparisons, supervision controls, and representation diagnostics
    assess overall utility and setting-dependent component effects.
\end{itemize}
\section{Related Work}
\label{sec:related_work}

\paragraph{Agent training and group-based credit.}
Agent RL learns from interaction outcomes
\citep{zhou2024archer,chen2025loop,wang2025ragen,luo2025agentlightning},
with GRPO, RLOO, and DAPO estimating within-group advantages
\citep{shao2024deepseekmath,ahmadian2024back,yu2025dapo}.
GiGPO groups anchor states; HGPO combines historical-context depths
\citep{feng2025gigpo,he2026hgpo}.
SALT matches recent interaction histories, while GraphGPO and RTMC aggregate
credit through transition graphs and shared-state trees
\citep{li2025salt,cheng2026graphgpo,wang2026rtmc}.
In the same sparse-reward, long-horizon setting, G$^2$PO merges identical
observations into a global state-transition graph and standardizes temporal-difference
errors across edges to assign action credit \citep{wang2026g2po}.
AT-GRPO adapts tree-expansion depth to dialogue stage \citep{peng2026atgrpo};
STAPO selects steps using state-normalized entropy and combines trajectory-aware
rewards with trajectory-independent penalties \citep{qi2026stapo}.
Other designs use peer baselines, progress signals, or hierarchical updates
\citep{wang2026bipace,fan2026progpo,yang2026progress,tan2026hcapo,jin2026himac}.
ECPO calibrates anchor-state credit using action-return shrinkage and reliability
gates \citep{li2026ecpo}.
\method{} instead constructs action-pair supervision: pre-action state and rubric
progress define comparability, two-action support selects the matching resolution,
and opposite-fold outcomes label an auxiliary pairwise objective.

\paragraph{Task-structured supervision.}
Task structure supports rubric rewards, exploration, and finite-state reward
representations \citep{gunjal2025rar,zhou2026ruscarl,icarte2022rewardmachines}.
RubricEM organizes stages and memory; GEAR models criterion dependencies
\citep{li2026rubricem,lv2026gear}.
SRaR, TRCA, and DRACO provide rubric-based step feedback
\citep{xie2026srar,zhang2026trca,gandhi2026draco}.
BEACON combines milestone groups with reward shaping \citep{wang2026beacon};
MileGPO infers milestones and traps from rollout graphs \citep{qian2026milegpo}.
\method{} uses task-specific progress extractors to condition comparisons,
with terminal outcomes supplying preference labels. SALT and MileGPO derive
their comparison structure from histories and observed graphs.

\paragraph{Process, preference, and value supervision.}
Process supervision extends terminal feedback
\citep{uesato2022process,lightman2024verify}.
DPO, REBEL, SimPO, and Contrastive Preference Learning optimize response
or sequential-policy comparisons
\citep{rafailov2023dpo,gao2024rebel,meng2024simpo,hejna2024cpl};
Step-DPO uses step preferences \citep{lai2024stepdpo}.
Agent Q, VinePPO, OmegaPRM, and RTPO obtain intermediate evidence through search
or additional continuations
\citep{putta2024agentq,kazemnejad2025vineppo,luo2024omegaprm,li2026rtpo}.
POISE and SAPO learn value baselines or joint policy--value models
\citep{choi2026poise,liang2026sapo}.
\method{} reuses collected sibling rollouts, separating candidate decisions
from label contributors with fixed folds. Appendix Table~\ref{tab:methods}
compares these designs.
\section{Methodology}
\label{sec:method}

\begin{figure}[!t]
\centering
\includegraphics[width=\linewidth]{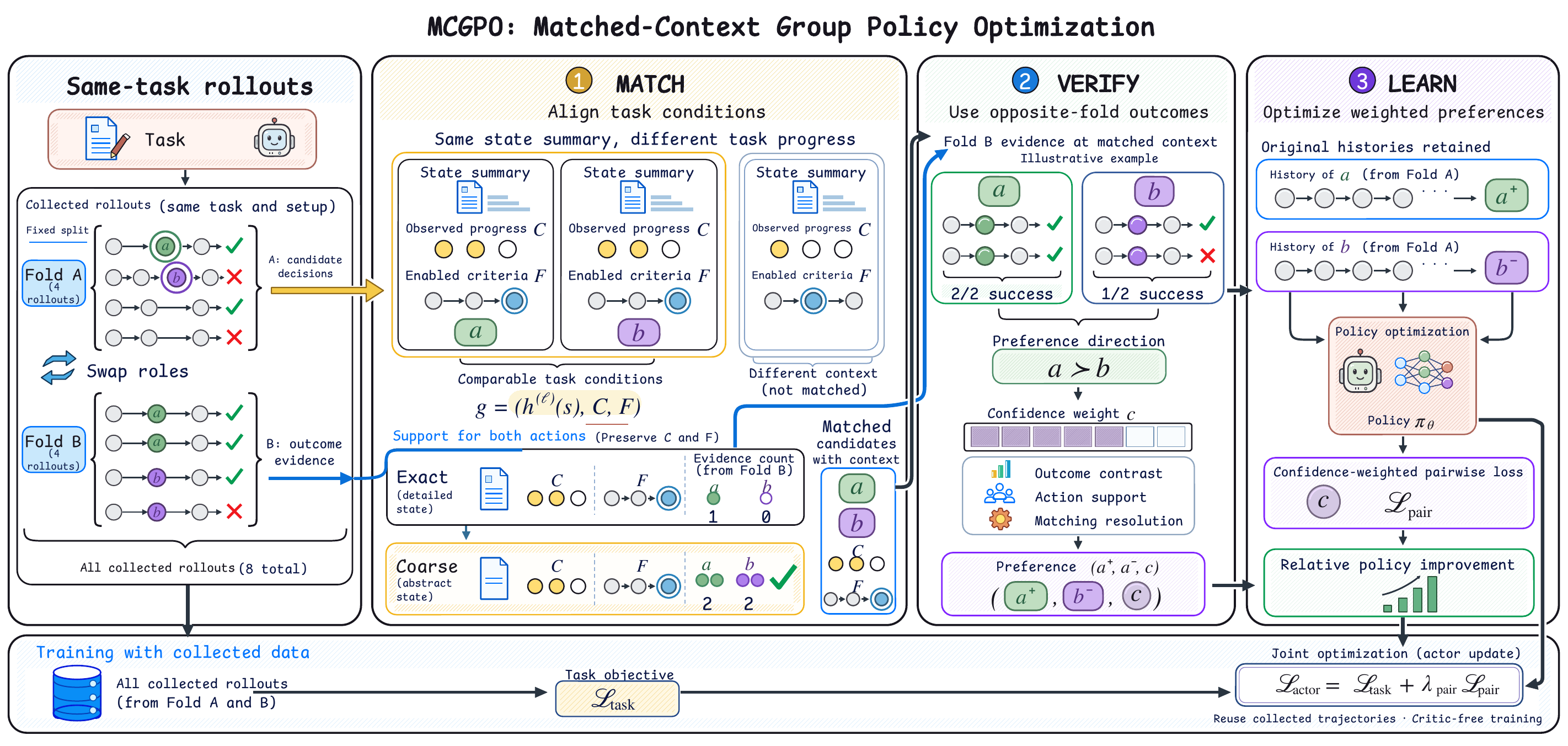}
\caption{Overview of \method{}. The Verify block labels matched actions
using opposite-fold terminal outcomes. The resulting evidence-weighted
pairwise loss complements the shared task objective.}
\label{fig:overview}
\end{figure}

For each task $x$, $\pi_{\rm old}$ collects eight sibling trajectories
$\tau_i$ containing observations $s_{i,t}$, responses $z_{i,t}$,
canonical actions $a_{i,t}$, and terminal success $y_i\in\{0,1\}$.
Rollout slots define fixed folds $\mathcal I_0,\mathcal I_1$ of four
trajectories each. \method{} turns these rollouts into local preferences
through \textbf{Match--Label--Learn} (Figure~\ref{fig:overview}). In each
direction, one fold supplies candidate decisions and the other supplies
outcome evidence; swapping roles uses both halves.

\subsection{Task-conditioned matching (Match)}

A comparison context combines the pre-action state summary with observed
task progress:
\begin{equation}
g_{i,t}^{(\ell)}=\big(h^{(\ell)}(s_{i,t}),C_{i,t},F_{i,t}\big),
\qquad \ell\in\{\mathrm{exact},\mathrm{coarse}\}.
\label{eq:frontier}
\end{equation}
Here $C_{i,t}$ contains rubric channels with finite positive pre-action
status, and $F_{i,t}$ annotates inactive criteria enabled by task rules.
Exact and coarse are ordered state-matching rules that preserve $(C,F)$;
coarse abstracts identity fields and can include task-relevant bins.
Appendix~\ref{app:matching-coordinates} specifies the fields and extractors.

At each rule, each trajectory contributes its first eligible visit to a
context. Candidate row pairs $p=((i,t),(j,u))\in\mathcal C_c^{(\ell)}$
come from distinct trajectories in fold $c$, share a key $g_p^{(\ell)}$,
and take distinct actions $a_p,b_p$. Let
$\mathcal E_{1-c}^{(\ell)}(g,a)$ contain opposite-fold trajectories whose
retained visit to $g$ takes action $a$, and let
$n_a^{(\ell)}(g)=|\mathcal E_{1-c}^{(\ell)}(g,a)|$.
The balanced support is
$n_\star^{(\ell)}(p)=\min\{n_{a_p}^{(\ell)}(g_p^{(\ell)}),
n_{b_p}^{(\ell)}(g_p^{(\ell)})\}$, giving
\begin{equation}
\ell^\star(p)=\min_{\mathrm{exact}\prec\mathrm{coarse}}
\{\ell:p\in\mathcal C_c^{(\ell)},\ n_\star^{(\ell)}(p)\ge m\}.
\label{eq:resolution-gate}
\end{equation}
We use $m=1$. Pairs without a supported rule are omitted.
This gate precedes the outcome test: a supported exact match retains
that rule even if its later contrast falls below threshold.
Appendix~\ref{app:comparison-groups} gives the first-visit and eligibility definitions.

\subsection{Cross-rollout preferences (Label)}

Let $\widehat p_{1-c}^{(\ell)}(g,a)$ be mean terminal success over
$\mathcal E_{1-c}^{(\ell)}(g,a)$. At the selected key and resolution,
\begin{align}
\widehat u_p
&=\widehat p_{1-c}^{(\ell^\star)}(g,a_p)
-\widehat p_{1-c}^{(\ell^\star)}(g,b_p),
\label{eq:uplift}\\
c_p&=|\widehat u_p|\,
\frac{n_\star}{n_\star+\kappa}\,q_{\ell^\star},
\qquad n_\star=n_\star^{(\ell^\star(p))}(p).
\label{eq:evidence-weight}
\end{align}
The sign of $\widehat u_p$ orders the rows as $(p^+,p^-)$; we retain
contrasts with $|\widehat u_p|\ge\epsilon$. The evidence weight combines
contrast magnitude, support shrinkage, and a preset resolution factor.
These labels express observational outcome contrasts under the rollout
policy; Appendix~\ref{app:label-provenance} details their conditional
outcome provenance.

For each sibling trajectory pair, we retain the earliest qualified
comparison, minimizing $\max(t_i,t_j)$ and breaking time ties by $c_p$.
A disjoint allocator keeps up to two preferences per fold, using each
selected decision row once. This yields a bounded preference set
$\mathcal P$ from the collected batch.
Algorithm~\ref{alg:mcgpo} and the parameter settings appear in
Appendix~\ref{app:algorithm}.

\subsection{Evidence-weighted policy learning (Learn)}

Each preferred response should receive a larger policy improvement than
its paired alternative. Using the original token histories $H_{i,t,k}$
and response mask $m_{i,t,k}$, define the length-normalized log-probability
change
\begin{equation}
d_\theta(i,t)=
\frac{\sum_k m_{i,t,k}
[\log\pi_\theta(z_{i,t,k}\mid H_{i,t,k})
-\log\pi_{\rm old}(z_{i,t,k}\mid H_{i,t,k})]}
{\max(1,\sum_k m_{i,t,k})}.
\label{eq:improvement}
\end{equation}
For pairs colocated in actor microbatch $\mathcal B$, the auxiliary loss is
\begin{equation}
\mathcal L_{\rm pair}=
\frac{\sum_{(+,-)\in\mathcal P_{\mathcal B}}
c_{+,-}\operatorname{softplus}(-[d_\theta(+)-d_\theta(-)])}
{\max(1,\lfloor|\mathcal B|/2\rfloor)}.
\label{eq:pair-loss}
\end{equation}
Normalization by the microbatch's pairing capacity makes the contribution
scale with supervision supply; $c_{+,-}$ controls each pair's influence.
The complete rollout group also supplies GraphGPO~\citep{cheng2026graphgpo}
task advantages, giving
\begin{equation}
\mathcal L_{\rm actor}(\mathcal B)=\mathcal L_{\rm task}(\mathcal B)
+\lambda_{\rm pair}\mathcal L_{\rm pair}.
\label{eq:actor}
\end{equation}
Here $\mathcal L_{\rm task}$ is the shared clipped policy loss with
reference-policy and entropy terms (Appendix~\ref{app:task-objective}).
Tasks without preferences still receive this task update. The construction
reuses sampled trajectories and policy log probabilities, preserving
critic-free training. Appendix~\ref{app:pair-gradient} derives the local
pairwise update.
\section{Experiments}
\label{sec:experiments}
\label{sec:results}

We organize the evaluation around four questions: whether \method{}
improves task completion and learning progress (RQ1), how comparison
conditions, outcome evidence, and preference learning contribute (RQ2),
how matched contexts produce weighted supervision on WebShop (RQ3), and whether
the WebShop gains extend to Qwen3 backbones (RQ4).

\subsection{Experimental Setup}
\label{sec:experimental-setup}

\subsubsection{Benchmarks and Models}
We evaluate household interaction in ALFWorld~\citep{shridhar2021alfworld}
and product search and purchase in WebShop~\citep{yao2022webshop}.
The main experiments use Qwen2.5-1.5B-Instruct and
Qwen2.5-7B-Instruct~\citep{yang2024qwen25}, with training seeds 0, 1,
and 2. GraphGPO supplies the shared task-credit objective and matched baseline,
measuring the effect of the complete rubric-conditioned preference branch.
WebShop rubric extraction uses structured training information from the
goal scorer, including attribute, option, and price matches and a
pre-purchase preview score (Appendix~\ref{app:matching-coordinates}).
Controls include GraphGPO with rubric-event shaping. For Qwen3-4B/8B,
\emph{GraphGPO + shaping} uses the same scorer information as \method{}
through training rewards, with three seeds per backbone.

\subsubsection{Implementation Details}
Within each backbone, MCGPO and GraphGPO use the same rollout, optimizer,
and environment settings. Following the agent protocol of
\citet{feng2025gigpo}, the Qwen2.5 policy conditions on the two most recent
interaction steps and generates reasoning inside \texttt{<think>} tags
followed by an action inside \texttt{<action>} tags. Each response is
limited to 512 generated tokens. The environment horizons are 50 turns
for ALFWorld and 15 for WebShop.

Each policy update uses 16 task groups with eight sibling rollouts per
group. We train for 250 policy updates with rollout temperature 1.0,
a constant learning rate of $10^{-6}$, weight decay 0.01, and KL
coefficient 0.01. The 1.5B and 7B runs use two and four GPUs,
respectively, with tensor-parallel sizes of two and four.
MCGPO uses the fixed preference-construction settings in
Table~\ref{tab:preference-settings}. The Qwen3 extension uses action-only
responses with thinking disabled. Resource settings, environment prompts,
and an illustrative rollout appear in Appendices~\ref{app:config},
\ref{app:prompts}, and~\ref{app:trace}, respectively.

\paragraph{Evaluation.}
Every five updates, we sample one rollout for each of the same 128
validation tasks at temperature 0.4. To reduce sensitivity to individual
checkpoints and sampled evaluation rollouts, our primary metric is mean
success over the last five scheduled evaluations ($L_5$). All matched
runs use the same end-of-budget window. We first average within
seed and then report the mean and sample SD across seeds.
Appendix~\ref{app:result-freezing} provides precise definitions,
seed-level results, and summaries of learning progress.

\subsection{RQ1: Does MCGPO Improve Task Completion and Learning Progress?}
\label{sec:rq1}

Table~\ref{tab:main-comparison} reports ALFWorld and WebShop results,
including published references, task-category success, and graded scores.
Appendix~\ref{app:result-freezing} (Table~\ref{tab:matched-readouts}) provides
paired uncertainty estimates and complementary readouts for the matched runs.

\begin{table}[!t]
\centering
\caption{Task success (\%) on ALFWorld and graded score/success (\%) on
WebShop. All and Succ. are the primary metrics.
GraphGPO and \method{} report mean\,$\pm$\,sample SD over three seeds
using the evaluation protocol in Section~\ref{sec:experimental-setup}.
Results for the other methods are from
\citet{feng2025gigpo,tan2026hcapo}.}
\label{tab:main-comparison}
\begingroup
\fontsize{7.5}{9}\selectfont
\setlength{\tabcolsep}{2.3pt}
\renewcommand{\arraystretch}{1.08}
\makeatletter
\newcommand{\metricsep}{%
  \begingroup\color{black!45}%
  \raisebox{-\dp\@arstrutbox}[0pt][0pt]{%
    \vbox to\dimexpr\ht\@arstrutbox+\dp\@arstrutbox\relax{%
      \offinterlineskip
      \cleaders\vbox{\hrule width0.35pt height1.4pt\kern1.4pt}\vfill}}%
  \endgroup}
\makeatother
\newcommand{\ms}[2]{\mbox{#1{\mdseries\fontsize{6.5}{7.2}\selectfont$\pm$#2}}}
\newcommand{\bms}[2]{\cellcolor{tablebest}\ms{\textbf{#1}}{#2}}
\newcommand{\ums}[2]{\cellcolor{tablesecond}\ms{\underline{#1}}{#2}}
\newcommand{\MCGPOLateFiveSmallRows}{%
\rowcolor{RCBaselineRow}RL & \graphgpo{}
& \ums{96.1}{1.2} & \ums{79.5}{4.7} & \ms{96.8}{2.8} & \ms{92.9}{4.1} & \ums{93.6}{1.6} & \bms{89.6}{5.4} & \ums{92.60}{1.93} & \bms{88.79}{1.26} & \ums{77.03}{1.22} \\
\rowcolor{RCMethodRow}RL & \textbf{\method{}}
& \bms{96.5}{4.2} & \bms{84.6}{0.9} & \ums{97.3}{2.0} & \bms{95.8}{2.9} & \bms{94.6}{3.2} & \ums{88.8}{2.7} & \bms{93.65}{1.40} & \ums{86.83}{3.29} & \bms{78.13}{2.97} \\
}

\newcommand{\MCGPOLateFiveLargeRows}{%
\rowcolor{RCBaselineRow}RL & \graphgpo{}
& \ums{99.2}{1.5} & \ums{91.6}{5.6} & \bms{99.8}{0.4} & \ums{90.7}{1.1} & \ms{90.2}{5.2} & \ums{92.2}{6.6} & \ums{94.84}{2.11} & \ums{90.23}{0.66} & \ums{80.73}{1.52} \\
\rowcolor{RCMethodRow}RL & \textbf{\method{}}
& \bms{99.8}{0.4} & \bms{98.6}{1.5} & \ms{98.3}{0.5} & \bms{93.1}{1.0} & \bms{94.0}{3.7} & \bms{92.3}{2.8} & \bms{96.30}{0.86} & \bms{90.28}{2.58} & \bms{81.20}{5.02} \\
}

\begin{tabular}{@{}ll*{6}{c}>{\bfseries}c!{\metricsep}c>{\bfseries}c@{}}
\toprule
\multirow{2}{*}{\textbf{Type}} & \multirow{2}{*}{\textbf{Method}}
& \multicolumn{7}{c!{\metricsep}}{\textbf{ALFWorld}}
& \multicolumn{2}{c}{\textbf{WebShop}} \\
\cmidrule(lr){3-9}\cmidrule(lr){10-11}
& & \textbf{Pick} & \textbf{Look} & \textbf{Clean} & \textbf{Heat}
& \textbf{Cool} & \textbf{Pick2}
& \cellcolor{tableaccent!75!black}\textcolor{white}{\textbf{All}}
& \textbf{Score}
& \cellcolor{tableaccent!75!black}\textcolor{white}{\textbf{Succ.}} \\
\midrule
\rowcolor{tablegroup}\multicolumn{11}{@{}l}{\textcolor{tableaccent}{\itshape Closed-source models}} \\
Prompting & GPT-4o
& 75.3 & 60.8 & 31.2 & 56.7 & 21.6 & 49.8 & 48.0 & 31.8 & 23.7 \\
Prompting & Gemini-2.5-Pro
& 92.8 & 63.3 & 62.1 & 69.0 & 26.6 & 58.7 & 60.3 & 42.5 & 35.9 \\

\specialrule{0.7pt}{2.2pt}{1.5pt}
\rowcolor{tablegroup}\multicolumn{11}{@{}l}{\textcolor{tableaccent}{\itshape Base: Qwen2.5-1.5B-Instruct}} \\
Prompting & Qwen2.5
& 5.9 & 5.5 & 3.3 & 9.7 & 4.2 & 0.0 & 4.1 & 23.1 & 5.2 \\
Prompting & ReAct
& 17.4 & 20.5 & 15.7 & 6.2 & 7.7 & 2.0 & 12.8 & 40.1 & 11.3 \\
Prompting & Reflexion
& 35.3 & 22.2 & 21.7 & 13.6 & 19.4 & 3.7 & 21.8 & 55.8 & 21.9 \\
RL & PPO
& \ms{64.8}{3.5} & \ms{40.5}{6.9} & \ms{57.1}{4.9}
& \ms{60.6}{6.6} & \ms{46.4}{4.0} & \ms{47.4}{1.9}
& \ms{54.4}{3.1} & \ms{73.8}{3.0} & \ms{51.5}{2.9} \\
RL & RLOO
& \ms{88.3}{3.0} & \ms{52.8}{8.6} & \ms{71.0}{5.9}
& \ms{62.8}{8.7} & \ms{66.4}{5.5} & \ms{56.9}{4.7}
& \ms{69.7}{2.5} & \ms{73.9}{5.6} & \ms{52.1}{6.7} \\
RL & GRPO
& \ms{85.3}{1.5} & \ms{53.7}{8.0} & \ms{84.5}{6.8}
& \ms{78.2}{7.9} & \ms{59.7}{5.0} & \ms{53.5}{5.6}
& \ms{72.8}{3.6} & \ms{75.8}{3.5} & \ms{56.8}{3.8} \\
RL & GiGPO$_{\rm w/std}$
& \ms{94.4}{5.9} & \ms{67.5}{4.6} & \ms{94.8}{3.8}
& \ums{94.4}{7.8} & \ms{79.8}{4.7} & \ms{76.4}{5.4}
& \ms{86.7}{1.7} & \ms{83.1}{1.6} & \ms{65.0}{3.2} \\
RL & GiGPO$_{\rm w/o\,std}$
& \ms{96.0}{1.4} & \ms{76.5}{3.9} & \ms{91.8}{5.5}
& \ms{91.3}{6.3} & \ms{71.7}{8.4} & \ms{79.5}{7.7}
& \ms{86.1}{4.7} & \ms{83.5}{1.8} & \ms{67.4}{4.5} \\
RL & HCAPO
& \ms{88.6}{7.0} & \ms{75.0}{0.0} & \bms{97.6}{1.8}
& \ms{90.7}{6.9} & \ms{84.2}{0.0} & \ms{74.2}{6.9}
& \ms{87.0}{4.1} & \ms{83.8}{0.7} & \ms{68.5}{1.0} \\
\MCGPOLateFiveSmallRows

\specialrule{0.7pt}{2.2pt}{1.5pt}
\rowcolor{tablegroup}\multicolumn{11}{@{}l}{\textcolor{tableaccent}{\itshape Base: Qwen2.5-7B-Instruct}} \\
Prompting & Qwen2.5
& 33.4 & 21.6 & 19.3 & 6.9 & 2.8 & 3.2 & 14.8 & 26.4 & 7.8 \\
Prompting & ReAct
& 48.5 & 35.4 & 34.3 & 13.2 & 18.2 & 17.6 & 31.2 & 46.2 & 19.5 \\
Prompting & Reflexion
& 62.0 & 41.6 & 44.9 & 30.9 & 36.3 & 23.8 & 42.7 & 58.1 & 28.8 \\
RL & PPO
& \ms{92.3}{4.0} & \ms{64.0}{8.4} & \ms{92.5}{2.4}
& \ms{89.5}{7.0} & \ms{80.3}{2.0} & \ms{68.8}{8.3}
& \ms{80.4}{2.7} & \ms{81.4}{3.1} & \ms{68.7}{5.1} \\
RL & RLOO
& \ms{87.6}{4.3} & \ms{78.2}{8.3} & \ms{87.3}{5.8}
& \ms{81.3}{7.6} & \ms{71.9}{5.2} & \ms{48.9}{8.4}
& \ms{75.5}{4.6} & \ms{80.3}{3.2} & \ms{65.7}{4.0} \\
RL & GRPO
& \ms{90.8}{5.1} & \ms{66.1}{6.7} & \ms{89.3}{5.4}
& \ms{74.7}{6.9} & \ms{72.5}{5.4} & \ms{64.7}{7.3}
& \ms{77.6}{5.2} & \ms{79.3}{2.8} & \ms{66.1}{3.7} \\
RL & GiGPO$_{\rm w/std}$
& \ms{97.7}{1.6} & \ms{82.7}{7.9} & \ums{98.8}{1.6}
& \ms{83.7}{7.2} & \ms{89.3}{8.2} & \ms{79.2}{6.6}
& \ms{90.8}{1.3} & \ms{84.4}{2.9} & \ms{72.8}{3.2} \\
RL & GiGPO$_{\rm w/o\,std}$
& \ms{91.8}{5.4} & \ms{88.6}{6.3} & \ms{95.9}{3.2}
& \ms{90.2}{2.6} & \ms{86.5}{5.5} & \ms{85.2}{7.5}
& \ms{90.2}{2.3} & \ms{86.2}{2.6} & \ms{75.2}{3.8} \\
RL & HCAPO
& \ms{99.1}{1.3} & \ms{90.3}{2.0} & \ms{97.3}{1.9}
& \ms{81.8}{8.8} & \ums{90.8}{6.6} & \ms{81.9}{10.0}
& \ms{91.4}{2.3} & \ms{85.1}{1.3} & \ms{73.8}{2.8} \\
\MCGPOLateFiveLargeRows
\bottomrule
\end{tabular}
\endgroup
\end{table}

\paragraph{Task completion.}
\method{} improves mean success over matched GraphGPO in
all four settings: by 1.04/1.46 percentage points on ALFWorld and
1.09/0.47 points on WebShop at 1.5B/7B. ALFWorld success rises from
92.60\%/94.84\% to 93.65\%/96.30\%, reducing the already-small
remaining failure rate by approximately 14\%/28\%.

\paragraph{Where the gains appear.}
ALFWorld's largest category gains occur on Look (5.07/6.93 points at
1.5B/7B); Heat and Cool also improve at both sizes. On WebShop, success
counts completed tasks, while the graded score also credits partial matches.
At 1.5B, higher success accompanies a graded-score decrease from 88.79
to 86.83; at 7B, the scores are nearly equal.

\begin{figure}[!htbp]
\centering
\includegraphics[width=\linewidth]{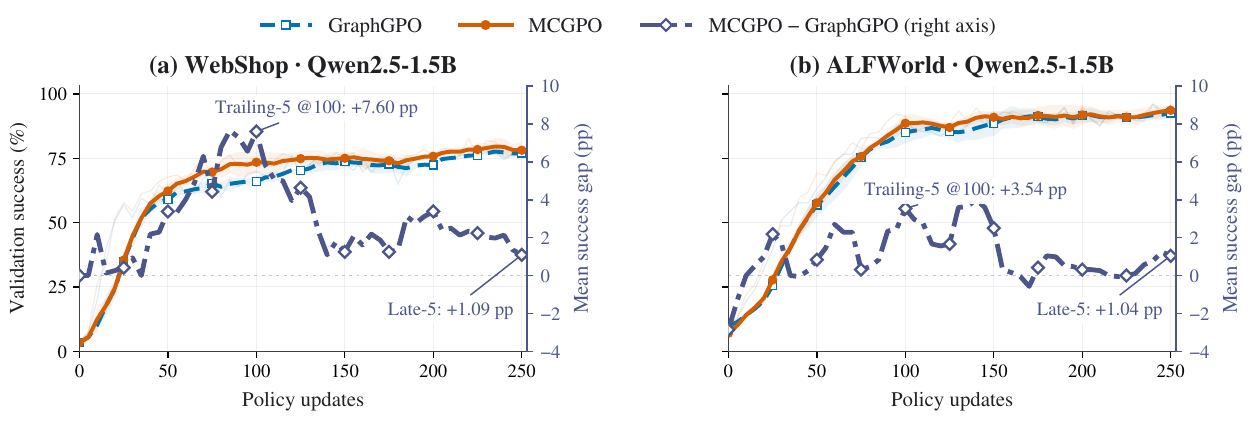}
\caption{Learning curves and paired gains with Qwen2.5-1.5B.
\textbf{Left axes:} validation success for \method{} and GraphGPO;
bands show the three-seed sample SD and faint lines retain raw means.
\textbf{Right axes:} the deep-blue curve shows the mean paired difference
(\method{} minus GraphGPO), in percentage points.
Each seed is averaged over up to five trailing evaluations, using the
available evaluations at the beginning so curves start at update 0;
complete five-point windows apply from update 20.
The update-100 annotation uses the corresponding smoothed values.
Shaded bands apply only to success curves.}
\label{fig:text-curves}
\end{figure}

\paragraph{Learning progress.}
Figure~\ref{fig:text-curves} shows an earlier rise in success for the
1.5B models. At update 100, the smoothed advantage is 7.60 points on
WebShop and 3.54 on ALFWorld. A retrospective summary of the complete,
unsmoothed curves integrates success over updates 0--250 and divides by
250 (Table~\ref{tab:matched-readouts}). The resulting gains are
1.25/1.46 points on ALFWorld and 2.98/$-0.02$ on WebShop at 1.5B/7B;
the first three settings improve in each of their three seeds.
This comparison measures learning progress along the policy-update axis.
We also compare task success at matched interaction budgets below.
Appendix~\ref{app:result-freezing} reports uncertainty
across the three training seeds;
Appendix~\ref{app:all-text-curves} gives all four curves.

\paragraph{Matched interaction budgets.}
We complement the update-axis curves with a retrospective budget comparison
on Qwen2.5-1.5B. For each training seed, we select the scheduled checkpoint pair with
the largest shared training-transition budget among pairs whose relative
budget difference is below 1\%. Selection uses budgets alone.
Figure~\ref{fig:matched-transition-budget-1p5b} in
Appendix~\ref{app:transition-budget-1p5b} reports raw checkpoint
success: \method{} raises the mean from 92.97\% to 95.57\% on ALFWorld
and from 76.04\% to 79.43\% on WebShop, gains of 2.60 and 3.39 points.
WebShop improves in all three paired seeds; ALFWorld's differences vary
across seeds. Appendix~\ref{app:transition-budget-1p5b} gives the selected
checkpoints, counted interactions, and uncertainty.

\subsection{RQ2: Which Parts of the Supervision Design Matter?}
\label{sec:rq2}

We test three design choices: task conditions for matching, the outcome
evidence used to orient preferences, and the objective that transfers
those preferences to the policy. Table~\ref{tab:mechanisms}(a) compares
Full with four Qwen2.5-1.5B variants, using one training run per variant
in each environment.

\begingroup
\setlength{\intextsep}{18pt plus 2pt minus 2pt}
\begin{table}[!tbp]
\centering
\small
\setlength{\tabcolsep}{5pt}
\renewcommand{\arraystretch}{1.08}
\setlength{\abovecaptionskip}{0pt}
\setlength{\belowcaptionskip}{6pt}
\caption{Component ablations of \method{}.
(a) One Qwen2.5-1.5B run per configuration; parentheses give decreases from Full (pp).
(b) Qwen3-4B/WebShop mean $\pm$ sample SD over three seeds.
Entries are success rates (\%) under the common evaluation protocol
(Section~\ref{sec:experimental-setup}); differences are computed before rounding.
Protocols and paired analyses: Appendix~\ref{app:mechanism-readouts}.}
\label{tab:mechanisms}
\begin{tabular*}{\linewidth}{@{\extracolsep{\fill}}llcc@{}}
\multicolumn{4}{@{}l}{\textbf{(a) Qwen2.5-1.5B: supervision controls (one run)}} \\[2pt]
\toprule
Component & Configuration & WebShop $\uparrow$ & ALFWorld $\uparrow$ \\
\midrule
\multicolumn{2}{@{}l}{\textbf{\method{} (Full)}} & \textbf{78.13} & \textbf{94.53} \\
\midrule
Match & w/ Shuffled Rubric & 75.78\enspace($\downarrow$2.34) & 93.91\enspace($\downarrow$0.63) \\
\addlinespace[2pt]
Label & Candidate-progress labels & 76.72\enspace($\downarrow$1.41) & 91.41\enspace($\downarrow$3.13) \\
\addlinespace[2pt]
\multirow{2}{*}[-1pt]{Learn} & w/o Pairwise Loss & 76.72\enspace($\downarrow$1.41) & 93.75\enspace($\downarrow$0.78) \\
\addlinespace[2pt]
 & GraphGPO + Rubric Shaping & 69.84\enspace($\downarrow$8.28) & 91.88\enspace($\downarrow$2.66) \\
\bottomrule
\end{tabular*}
\par\vspace{1.2em}
\begin{tabular*}{\linewidth}{@{\extracolsep{\fill}}lccc@{}}
\multicolumn{4}{@{}l}{\textbf{(b) Qwen3-4B / WebShop: $2\times2$ controls (three seeds)}} \\[2pt]
\toprule
Configuration & Opposite fold & Evidence weights & Success (\%) $\uparrow$ \\
\midrule
A: \method{} (Full) & Yes & Yes & $\mathbf{82.14} \pm 3.52$ \\
B: Same-fold terminal & No & Yes & $79.79 \pm 3.37$ \\
C: Uniform weighting & Yes & No & $80.52 \pm 3.75$ \\
D: Same-fold + uniform & No & No & $78.13 \pm 3.36$ \\
\bottomrule
\end{tabular*}

\end{table}
\endgroup

\paragraph{Comparison conditions and outcome evidence.}
\emph{w/ Shuffled Rubric} permutes pre-action $(C,F)$ labels among
same-task decision rows, disrupting their alignment with actual progress.
Success falls by 2.34 points on WebShop and 0.63 on ALFWorld.
The \emph{Candidate-progress labels} variant uses the candidate
trajectories' progress evidence in place of opposite-fold terminal
outcomes to label comparisons, reducing success by 1.41 and 3.13 points. This joint replacement
changes evidence source, label content, and available support. Its result
favors the complete terminal-outcome preference design over the
candidate-progress alternative.

\paragraph{How the supervision enters training.}
\emph{w/o Pairwise Loss} sets $\lambda_{\rm pair}=0$ while retaining
preference diagnostics; its success drops by 1.41 and 0.78 points.
\emph{GraphGPO + Rubric Shaping} adds positive-progress rewards and
regression penalties to the task rewards, retaining GraphGPO credit
assignment with the pairwise loss disabled. This event-shaping baseline
scores 8.28 and 2.66 points below Full. It compares two uses of the rubric
channels under the stated shaping rule and coefficient.
The three-seed Qwen3-4B/WebShop factorial controls in
Table~\ref{tab:mechanisms}(b) show gains of
\factorFoldWeightedGain{}--\factorFoldUniformGain{} points from opposite-fold
evidence and \factorWeightOppositeGain{}--\factorWeightSameGain{} points from
evidence weighting, with all four contrasts positive in every seed.
Appendix~\ref{app:factorial-controls} reports configurations and paired intervals.
The Qwen3 experiments additionally compare \method{} with GraphGPO + shaping
under the same scorer access (Appendix~\ref{app:scorer-matched-control}).

\paragraph{Comparison representation.}
Joint state--rubric conditioning gives the highest three-seed mean
success in all four settings, exceeding State-only by
\repStateMin{}--\repStateMax{} percentage points and Rubric-only by
\repRubricMin{}--\repRubricMax{} points
(Appendix~\ref{app:representation-controls}). The gains persist at the
terminal evaluation in all four settings. With preference construction
and the learning objective fixed, these gains support conditioning action
comparisons and their terminal evidence jointly on state and progress.
WebShop State-only already retains task-match score bins in its coarse
state, placing this comparison within a scorer-assisted training setting.

\subsection{RQ3: What Preferences and Weights Does the Construction Supply?}
\label{sec:rq3}

\begin{figure}[!htbp]
\centering
\includegraphics[width=\linewidth]{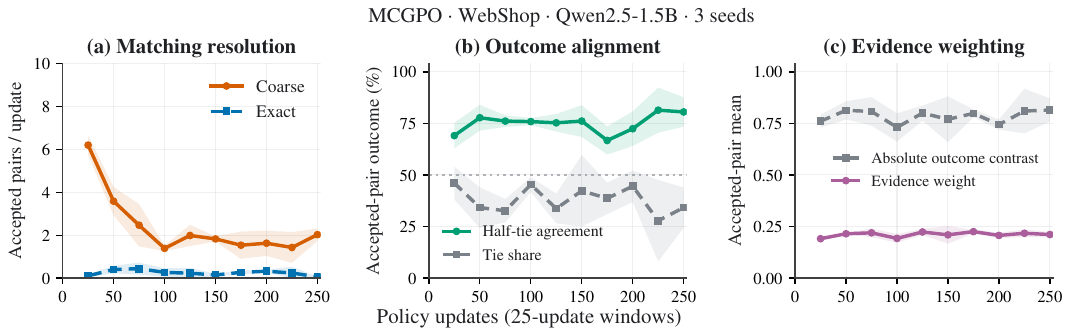}
\caption{Preference construction on WebShop with Qwen2.5-1.5B.
\textbf{(a)} Accepted pairs from exact and coarse matching.
\textbf{(b)} Agreement of preference directions with candidate outcomes
(ties receive half credit), together with the tie share.
\textbf{(c)} Absolute outcome contrast and the resulting evidence weight,
averaged over accepted pairs.
Each point summarizes 25 updates; lines and bands show the three-seed
mean and $\pm1$ sample SD. Ratios and pair-level means pool observations
within each seed/window. Appendix~\ref{app:all-text-curves} defines the
statistics and aggregation.}
\label{fig:preference-dynamics}
\end{figure}

We examine preference construction on WebShop with Qwen2.5-1.5B,
following supported comparisons through outcome-based labeling and
evidence-weighted learning. Figure~\ref{fig:preference-dynamics} covers
all 750 updates from three runs: the matching resolution
that supplies pairs, agreement with candidate outcomes, and the weight
assigned to the evidence.

\paragraph{Matching supplies comparisons under sparse action support.}
Of 1,990 accepted preferences, 1,805 (90.70\%) use coarse matching,
which relaxes state detail while retaining rubric progress and dependency
coordinates. Average supply declines from 3.99 pairs per update in
updates 26--50 to 2.08 in 226--250, with coarse matching dominant throughout.
The weaker-supported action has one evidence trajectory in 1,547 pairs
(77.74\%); the remaining 443 have two for each action.

\paragraph{Outcome evidence determines direction and strength.}
On WebShop, the preferred candidate succeeds while its alternative fails in 1,093
pairs; the reverse occurs in 124, and 773 have equal outcomes.
This gives 74.35\% pooled agreement when ties receive half credit.
Every plotted window has a three-seed mean above 50\%, showing a
persistent association between preferences inferred from the evidence
fold and outcomes in the separate candidate fold.
The evidence fold's mean absolute outcome contrast is 0.781; accounting
for action support and matching resolution yields a mean evidence
weight of 0.208 (Figure~\ref{fig:preference-dynamics}c).
Agreement measures association with candidate terminal outcomes,
which also reflect subsequent actions.

\paragraph{Connection to policy learning.}
The auxiliary loss is active in all 511 updates with accepted preferences
(68.13\% of training updates). RQ2 complements these supervision diagnostics
with loss-removal, shuffled-rubric, and representation controls of policy
performance.

\subsection{RQ4: Do the WebShop Gains Extend to Qwen3 Backbones?}
\label{sec:rq4}

We extend the matched GraphGPO--\method{} comparison to Qwen3-4B and
8B on WebShop with seeds 0, 1, and 2. These runs share the training
budget and evaluation schedule and use action-only responses with
thinking disabled.

\begin{table}[!htbp]
\centering
\small
\setlength{\tabcolsep}{8pt}
\renewcommand{\arraystretch}{1.08}
\setlength{\abovecaptionskip}{0pt}
\setlength{\belowcaptionskip}{6pt}
\caption{Task success on WebShop with Qwen3 backbones. Entries are
mean $\pm$ sample SD over three training seeds under the evaluation
protocol in Section~\ref{sec:experimental-setup}. Gain is the paired
mean difference in percentage points, computed before rounding.}
\label{tab:qwen3-transfer}
\begin{tabular}{@{}lrrr@{}}
\toprule
\multirow{2}{*}{Backbone} & \multicolumn{2}{c}{Mean success (\%)} & \multirow{2}{*}{Gain (pp)} \\
\cmidrule(lr){2-3}
& GraphGPO & \method{} & \\
\midrule
Qwen3-4B & $78.18 \pm 2.62$ & $\mathbf{82.14} \pm 3.52$ & +3.96 \\
Qwen3-8B & $79.69 \pm 1.95$ & $\mathbf{84.84} \pm 1.96$ & +5.16 \\
\bottomrule
\end{tabular}
\end{table}

\method{} raises mean success from \qwenFourBase{}\% to \qwenFourOurs{}\%
at 4B and from \qwenEightBase{}\% to \qwenEightOurs{}\% at 8B, gains of
\qwenFourGain{} and \qwenEightGain{} percentage points, respectively.
All six paired runs improve, with gains of 2.66--5.78 points at 4B and
2.50--9.22 points at 8B (Table~\ref{tab:qwen3-seeds}). This consistency
across seeds extends the evidence to both Qwen3 model sizes under
action-only responses.
A scorer-matched reward-shaping comparison appears in
Appendix~\ref{app:scorer-matched-control}.
\section{Conclusion}

\method{} constructs action preferences by matching state and rubric progress
within existing rollout groups. Opposite-fold terminal outcomes supply
preference directions, and an evidence-weighted pairwise loss complements
the GraphGPO task objective.

Across six environment--backbone settings, \method{} improves three-seed
mean success over matched GraphGPO, including ALFWorld baselines above
92\% and Qwen3 WebShop gains of \qwenFourGain{} and \qwenEightGain{} points.
Representation and factorial controls support joint context matching,
opposite-fold evidence, and evidence weighting; scorer-matched comparisons
favor \method{} over reward shaping. Together, these results support
extracting useful action-level supervision from existing rollout groups.

\subsection*{AI Use Statement}

Generative AI tools assisted with code development, experiment-log
organization and analysis, manuscript drafting, and language editing.
The authors are responsible for independently verifying all algorithmic
claims, references, experimental results, and final manuscript content.

\bibliography{iclr2027_conference}
\bibliographystyle{iclr2027_conference}
\clearpage
\appendix
\raggedbottom
\setcounter{topnumber}{4}
\setcounter{bottomnumber}{3}
\setcounter{totalnumber}{6}
\renewcommand{\topfraction}{0.92}
\renewcommand{\bottomfraction}{0.85}
\renewcommand{\textfraction}{0.06}
\renewcommand{\floatpagefraction}{0.90}
\setlength{\parskip}{2pt plus 1pt minus 1pt}
\setlength{\textfloatsep}{8pt plus 2pt minus 2pt}
\setlength{\floatsep}{6pt plus 2pt minus 2pt}
\setlength{\intextsep}{6pt plus 2pt minus 2pt}
\setlength{\abovecaptionskip}{6pt}
\makeatletter
\setlength{\@fptop}{0pt}
\setlength{\@fpsep}{10pt plus 2pt minus 2pt}
\setlength{\@fpbot}{0pt plus 1fil}
\makeatother

\section{Algorithm and comparison construction}
\label{app:algorithm}

This section specifies the comparison sets, evidence weights, and actor
objective used in Section~\ref{sec:method}. Algorithm~\ref{alg:mcgpo}
summarizes the complete update. The same eight sibling trajectories
supply the task objective and the auxiliary preferences.

\begin{algorithm}[H]
\caption{MCGPO: Match--Label--Learn}
\label{alg:mcgpo}
\begingroup
\small
\renewcommand{\algorithmicrequire}{\textbf{Require:}}
\renewcommand{\algorithmiccomment}[1]{\hfill{\color{RCBlue}$\triangleright$~#1}}
\begin{algorithmic}[1]
\REQUIRE Policy $\pi_\theta$, tasks $\mathcal D$, support $m$,
contrast threshold $\epsilon$, shrinkage $\kappa$, pair weight $\lambda_{\rm pair}$.
\FOR{each training iteration}
  \STATE $\pi_{\rm old}\gets\pi_\theta$;
  $\mathcal I_0\gets\{0,2,4,6\}$, $\mathcal I_1\gets\{1,3,5,7\}$
  \STATE $\mathcal T\gets\textsc{Collect}(\pi_{\rm old},\mathcal D,8)$;
  $\mathcal P\gets\emptyset$
  \STATE $A^{\rm task}\gets\textsc{TaskCredit}(\mathcal T)$
  \FOR{each admitted task $x$ and candidate fold $c\in\{0,1\}$}
    \STATE $g_{i,t}^{(\ell)}\gets(h^{(\ell)}(s_{i,t}),C_{i,t},F_{i,t})$
    \COMMENT{Match}
    \STATE $\mathcal C_c\gets\textsc{FirstSupportedMatch}_{\rm exact\to coarse}
      (\mathcal I_c,\mathcal I_{1-c};g,m)$
    \FOR{each matched decision-row pair $p\in\mathcal C_c$ with context $g$, actions $a,b$, and level $\ell$}
      \STATE $\widehat u_p\gets\widehat p_{1-c}^{(\ell)}(g,a)
        -\widehat p_{1-c}^{(\ell)}(g,b)$ \COMMENT{Label}
      \STATE $n_p\gets\min(n_a,n_b)$;
      $c_p\gets|\widehat u_p|\,n_p/(n_p+\kappa)\,q_\ell$
    \ENDFOR
    \STATE $\mathcal V_c\gets\{(\textsc{Orient}(p,\operatorname{sign}\widehat u_p),c_p):
       p\in\mathcal C_c,\ |\widehat u_p|\geq\epsilon\}$
    \STATE $\mathcal P\gets\mathcal P\cup
      \textsc{Disjoint}_2(\textsc{EarliestQualified}(\mathcal V_c))$
  \ENDFOR
  \FOR{each pair-aware actor minibatch $\mathcal M$}
    \STATE Compute $d_\theta$ and $\mathcal L_{\rm pair}$ on its microbatches
      using Eqs.~\ref{eq:improvement}--\ref{eq:pair-loss} \COMMENT{Learn}
    \STATE Update $\theta$ with $\mathcal L_{\rm task}
      +\lambda_{\rm pair}\mathcal L_{\rm pair}$ using $A^{\rm task}$ and $\mathcal P$
  \ENDFOR
\ENDFOR
\end{algorithmic}
\endgroup
\end{algorithm}

Table~\ref{tab:methods} compares the task and preference interfaces
with the related designs discussed in the main text.
\begin{table}[!htbp]
\centering
\caption{Comparison conditions, evidence, and policy interfaces.
Rows summarize method designs; the matched performance evaluation in
this paper uses GraphGPO. MCGPO's frontier is a rule-derived annotation
of progress.}
\label{tab:methods}
\footnotesize
\setlength{\tabcolsep}{4pt}
\renewcommand{\arraystretch}{1.10}
\begin{tabular}{@{}>{\raggedright\arraybackslash}p{0.15\linewidth}
>{\raggedright\arraybackslash}p{0.25\linewidth}
>{\raggedright\arraybackslash}p{0.34\linewidth}
>{\raggedright\arraybackslash}p{0.19\linewidth}@{}}
\toprule
\textbf{Method} & \textbf{Comparison conditions}
& \textbf{Evidence and aggregation} & \textbf{Policy interface} \\
\midrule
GiGPO \citep{feng2025gigpo} & Repeated anchor states
& Returns within trajectory and state groups & Trajectory and step advantages \\
HGPO \citep{he2026hgpo} & Current state and historical-context hierarchy
& Weighted advantages across context depths & Step advantage \\
BiPACE \citep{wang2026bipace} & Behavioral clusters from policy features
& Action-conditioned peer returns & Step advantage \\
\graphgpo{} \citep{cheng2026graphgpo} & Observed transition graph
& Changes in distance to observed goals & Graph advantage \\
G$^2$PO \citep{wang2026g2po} & Identical observations in a global transition graph
& Graph state values and globally standardized edge TD errors & Edge and episode advantages \\
AT-GRPO \citep{peng2026atgrpo} & Dialogue trees with adaptive expansion depth
& Returns over stage-dependent observation ranges & Group-relative policy loss \\
STAPO \citep{qi2026stapo} & Anchor states and normalized-entropy outliers
& Trajectory-aware rewards and trajectory-independent penalties & Selective step advantages \\
GEAR \citep{lv2026gear} & Dependencies among response criteria
& Dependency-adjusted criterion probabilities & Scalar rubric reward \\
\specialrule{0.7pt}{1.8pt}{1.0pt}
\textbf{\method{}} & State and observed progress with rule annotation
& Two-action support selects resolution; opposite-fold outcomes label pairs
& Auxiliary weighted pairwise objective \\
\bottomrule
\end{tabular}
\end{table}

\subsection{Comparison keys and first visits}
\label{app:matching-coordinates}
\label{app:comparison-groups}

Each decision has a pre-action state summary $h$, active rubric criteria
$C$, and rule annotation $F$. Boolean true and finite positive channel
values activate membership in $C$; extractor diagnostics are excluded.
Appendix~\ref{app:rubric-rules} gives the environment-specific channels.
For vocabulary $V$, prerequisites $P(j)$, blocking sources $B(j)$, and
the set $V_{\rm block}$ of blocking-source nodes,
\begin{equation}
F(C;V,G_x)=\{j\in V\setminus(C\cup V_{\rm block}):
P(j)\subseteq C,\ B(j)\cap C=\emptyset\}.
\label{eq:frontier-rules}
\end{equation}
$F$ annotates inactive criteria whose prerequisites are satisfied and
which are not blocked. The opening status is empty; later records use the
extractor schema. In ALFWorld, this vocabulary change can affect the
first evidence row retained at the opening boundary.

\begin{table}[!htbp]
\centering\small
\setlength{\tabcolsep}{4pt}
\caption{State fields for the ordered matching rules. Both rules retain
the rubric signature $(C,F)$.}
\label{tab:matching-fields}
\begin{tabular}{@{}p{0.13\linewidth}p{0.40\linewidth}p{0.41\linewidth}@{}}
\toprule
Environment & Exact state & Coarse state \\
\midrule
ALFWorld & Task type, location, held object, acquired/processed/placed
object sets, light status
& Task type, location, holding/processing flags, placement count,
light status, remaining-budget bin \\
WebShop & Page, search query, product identity, selected options
& Page, searched/product-open flags, binned type/attribute/option match,
price satisfaction \\
\bottomrule
\end{tabular}
\end{table}

Rollout slots define $\mathcal I_0=\{0,2,4,6\}$ and
$\mathcal I_1=\{1,3,5,7\}$. For key $g$, each trajectory contributes its
first eligible visit:
\begin{equation}
\begin{aligned}
t_i^{(\ell)}(g;b)&=\min\{t\ge b:g_{i,t}^{(\ell)}=g\},\\
\mathcal G_c^{(\ell)}(g;b)
&=\{(i,t_i^{(\ell)}(g;b)):i\in\mathcal I_c,\ t_i^{(\ell)}(g;b)<\infty\}.
\end{aligned}
\label{eq:comparison-groups}
\end{equation}
Candidate groups use $b=1$, while evidence groups use $b=0$.
Candidates have different canonical actions. Opposite-fold first visits
provide the support counts and empirical action-success rates used by
the support gate in Eq.~\ref{eq:resolution-gate}. The constructor retains
the earliest qualified comparison per trajectory pair, then allocates
up to two disjoint decision-row pairs per task and direction.

\begin{table}[!htbp]
\centering\small
\caption{Fixed preference-construction and learning settings.}
\label{tab:preference-settings}
\begin{tabular}{@{}lr@{}}
\toprule
Setting & Value \\
\midrule
Sibling trajectories / fold sizes & $8$ / $4+4$ \\
Minimum support for each action $m$ & $1$ \\
Terminal-score threshold for evidence & $0.999$ \\
Outcome-contrast threshold $\epsilon$ & $0.10$ \\
Support shrinkage $\kappa$ & $2$ \\
Resolution weights $(q_{\rm exact},q_{\rm coarse})$ & $(1,0.70)$ \\
Accepted pairs per task and direction & at most $2$ \\
Auxiliary-loss coefficient $\lambda_{\rm pair}$ & $0.05$ \\
\bottomrule
\end{tabular}
\end{table}

The four-trajectory evidence fold permits balanced action support
$n_\star\in\{1,2\}$. Its smallest nonzero binary-outcome contrast is
$1/3$, so the configured threshold retains every non-tied supported
contrast. Shrinkage contributes $1/3$ or $1/2$ at the two support levels.
Invalid groups keep their task update with zero auxiliary weight.

\subsection{Outcome provenance and pairwise update}
\label{app:label-provenance}

Let $D$ fix the admitted batch, rollout slots, keys, actions, turn indices,
and hyperparameters. In candidate direction $c$,
\begin{equation}
\mathcal P_c(D;y_c,y_{1-c})=\Phi_c(D,y_{1-c}).
\label{eq:label-provenance}
\end{equation}
Thus changing $y_c$ leaves that direction's labels and weights unchanged
under these fixed inputs. The reverse direction uses $y_c$ as evidence.
Batch admission uses whole-group outcomes and lies outside this
conditional property. Finite-support contrasts remain observational
estimates; the evidence weight controls their optimization contribution.

\paragraph{Local pairwise gradient.}
\label{app:pair-gradient}
During an actor update, pair membership, orientation, the rollout policy,
and weight $c_p$ are fixed. With
$\delta_\theta=d_\theta(+)-d_\theta(-)$ and
$g_\theta^\pm=\nabla_\theta d_\theta(\pm)$, one pair contributes
\begin{equation}
\nabla_\theta\ell_p
=-c_p\,\sigma(-\delta_\theta)(g_\theta^+-g_\theta^-).
\label{eq:pair-gradient}
\end{equation}
The actor colocates both rows in a microbatch, applies the normalization
in Eq.~\ref{eq:pair-loss}, and adds the scaled auxiliary term to the task
loss. RQ2 measures the policy effect of this addition.

\subsection{Shared task objective}
\label{app:task-objective}

The GraphGPO task branch uses
$R_{i,t}^{\rm graph}=10\gamma^{D(s_{i,t+1})}$, where $D$ is directed
distance to an observed successful goal; unreachable states receive one
more than the largest finite distance. Let $Z_{i,t}=\sum_k r^{\rm tok}_{i,t,k}$
be the token-level task score, including invalid-action penalties.
The advantage sums the normalized graph return within the task's anchor
state group and the normalized task score within its response group:
\begin{equation}
\widehat A_{i,t}^{\rm task}
=\operatorname{Norm}_{\mathcal U_{x,s_{i,t}}}(R_{i,t}^{\rm graph})
+\operatorname{Norm}_{\mathcal U_x}(Z_{i,t}).
\label{eq:task-advantage}
\end{equation}
Normalization uses the group mean and sample SD with $10^{-6}$ in the
denominator. Singleton step groups have centered score zero; singleton
task groups use mean zero and unit scale. Both task-credit coefficients
are one, and each row advantage is broadcast to its valid response tokens.

For token ratio $\rho=\pi_\theta/\pi_{\rm old}$, define
$\ell_{\rm clip}=\max\{-\rho A,
-\operatorname{clip}(\rho,0.8,1.2)A\}$ and
$\ell_{\rm dc}=\min\{\ell_{\rm clip},-3A\}$ when $A<0$,
with $\ell_{\rm dc}=\ell_{\rm clip}$ otherwise. The masked token mean
$\langle\cdot\rangle_{\mathcal B,m}$ gives
\begin{equation}
\mathcal L_{\rm task}
=\left\langle
\ell_{\rm dc}(\rho,\widehat A^{\rm task})
+0.01\,\widehat K-0.001\,\mathcal H(\pi_\theta)
\right\rangle_{\mathcal B,m},
\label{eq:task-loss}
\end{equation}
where $\widehat K=\operatorname{clip}(e^v-v-1,-10,10)$ and
$v=\log\pi_{\rm ref}-\log\pi_\theta$ on sampled tokens.
The mask follows the response format in Appendix~\ref{app:prompts}.
Gradient accumulation and distributed reduction use the shared actor
configuration.
\section{Experimental setup}
\label{app:config}

This section records the shared training recipe used by GraphGPO and
\method{}. The environment prompts appear in Appendix~\ref{app:prompts}.
Comparisons within a
backbone use the same task data, response protocol, optimizer, and
evaluation grid.

\subsection{Training and evaluation configuration}
Table~\ref{tab:setup} summarizes the environment and backbone profiles.
Each update uses $P\!\times\!G$ admitted trajectories; shared dynamic
group filtering can require additional generated groups when enabled.
\begin{table}[!htbp]
\caption{Training and rollout configuration for matched GraphGPO and
\method{} runs. Micro-batch sizes are reported per GPU. Mean--std denotes
the shared GraphGPO within-group advantage normalization mode.
TP denotes rollout tensor parallelism.}
\label{tab:setup}
\label{tab:resources}
\centering
\small\rmfamily
\setlength{\tabcolsep}{4.3pt}
\renewcommand{\arraystretch}{1.08}
\begin{tabular}{@{}lcccc@{}}
\toprule
 & \multicolumn{2}{c}{ALFWorld} & \multicolumn{2}{c}{WebShop} \\
\cmidrule(lr){2-3}\cmidrule(l){4-5}
Setting & Small & Large & Small & Large \\
\midrule
Qwen2.5 backbone & 1.5B & 7B & 1.5B & 7B \\
Qwen3 backbone & --- & --- & 4B & 8B \\
\midrule
Task groups / update ($P$) & 16 & 16 & 16 & 16 \\
Trajectories / group ($G$) & 8 & 8 & 8 & 8 \\
Admitted trajectories / update & 128 & 128 & 128 & 128 \\
Training updates & 250 & 250 & 250 & 250 \\
Max. environment steps & 50 & 50 & 15 & 15 \\
Observation history (pairs) & 2 & 2 & 2 & 2 \\
Max. prompt length (tokens) & 2048 & 2048 & 5120 & 5120 \\
Max. response length (tokens) & 512 & 512 & 512 & 512 \\
\midrule
PPO mini-batch & 256 & 64 & 64 & 64 \\
Actor micro-batch & 32 & 4 & 8 & 4 \\
Log-prob. micro-batch & 32 & 16 & 16 & 16 \\
GPUs & 2 & 4 & 2 & 4 \\
TP & 2 & 4 & 2 & 4 \\
\midrule
GraphGPO normalization & Mean--std & Mean--std & Mean--std & Mean--std \\
Discount $\gamma$ & 0.10 & 0.10 & 0.20 & 0.20 \\
\bottomrule
\end{tabular}
\end{table}

Every setting trains for 250 policy updates with eight sibling rollouts
per task, response length at most 512 tokens, rollout temperature 1.0,
learning rate $10^{-6}$ with a constant schedule, and weight decay 0.01.
Gradient checkpointing is enabled. The task objective's clipping and
regularization coefficients are specified in Appendix~\ref{app:task-objective}.
Validation samples one rollout on each of the same 128 tasks every five
updates at temperature 0.4.

\paragraph{Continuation and evaluation lineage.}
Main runs retain their archived $0\!\to\!150\!\to\!250$ protocol.
Mechanism controls target 250 updates from initialization and use their
corresponding Full reference seed (WebShop seed 0; ALFWorld seed 1).
The ALFWorld controls use the corrected epoch-loop budget.
Recovery restores model, optimizer, scheduler, random-number state,
dataloader state, and the update counter. Reported scheduled evaluations
follow the retained checkpoint lineage; reload evaluations and
rolled-back branches are excluded. Controls are separate executions, so
their recorded recovery histories can differ.
\section{Supplementary policy results and ablations}
\label{app:result-freezing}

These results complement RQ1, RQ2, and RQ4 with seed-level readouts,
uncertainty, and the control definitions. Success is reported in percent;
paired gains use percentage points.

\subsection{Readouts and seed-level results}
Our primary metric summarizes late-training success over the final five
scheduled evaluations of the shared 250-update budget. Averaging these
evaluations reduces sensitivity to an individual checkpoint and its
sampled rollouts. For each training seed, we compute
\begin{equation}
L_5=\frac{1}{5}\sum_{t\in\{230,235,240,245,250\}}S(t).
\label{eq:late5}
\end{equation}
Here $S(t)$ is validation success at update $t$. The same budget and
end-of-budget window apply to all matched methods and seeds.
We average within each seed before computing the mean and sample SD
across seeds. Table~\ref{tab:text-seeds} retains all designated Qwen2.5
seeds; Table~\ref{tab:qwen3-seeds}
provides the three-seed WebShop extension underlying
Table~\ref{tab:qwen3-transfer}.

\begin{table}[!htbp]
\centering\small
\caption{Per-seed Qwen2.5 policy success (\%). $L_5$ averages the final five scheduled evaluations of the 250-update budget; $E_{250}$ reports the final evaluation.}
\label{tab:text-seeds}
\setlength{\tabcolsep}{4pt}
\begin{tabular}{@{}llrrrr@{}}
\toprule
Setting & Seed & \method{} $L_5$ & GraphGPO $L_5$ & \method{} $E_{250}$ & GraphGPO $E_{250}$ \\
\midrule
ALFWorld / 1.5B & 0 & 92.03 & 90.47 & 97.66 & 92.97 \\
 & 1 & 94.53 & 93.13 & 94.53 & 96.88 \\
 & 2 & 94.38 & 94.22 & 94.53 & 96.88 \\
\midrule
ALFWorld / 7B & 0 & 95.31 & 96.88 & 96.88 & 95.31 \\
 & 1 & 96.88 & 95.00 & 96.88 & 96.88 \\
 & 2 & 96.72 & 92.66 & 94.53 & 92.97 \\
\midrule
WebShop / 1.5B & 0 & 78.13 & 76.25 & 79.69 & 75.00 \\
 & 1 & 81.09 & 78.44 & 78.91 & 77.34 \\
 & 2 & 75.16 & 76.41 & 67.19 & 75.78 \\
\midrule
WebShop / 7B & 0 & 84.84 & 79.06 & 86.72 & 83.59 \\
 & 1 & 83.28 & 81.09 & 81.25 & 78.13 \\
 & 2 & 75.47 & 82.03 & 72.66 & 78.91 \\
\bottomrule
\end{tabular}
\end{table}

\begin{table}[!htbp]
\centering\small
\caption{All paired Qwen3 WebShop seeds for the primary $L_5$ readout.
Each entry averages updates 230, 235, 240, 245, and 250.
Gains are computed before final display rounding; three-seed means appear in
Table~\ref{tab:qwen3-transfer}.}
\label{tab:qwen3-seeds}
\begin{tabular}{@{}llrrr@{}}
\toprule
Backbone & Seed & GraphGPO (\%) & \method{} (\%) & Gain (pp) \\
\midrule
Qwen3-4B & 0 & 79.84 & 85.63 & +5.78 \\
 & 1 & 75.16 & 78.59 & +3.43 \\
 & 2 & 79.53 & 82.19 & +2.66 \\
\midrule
Qwen3-8B & 0 & 81.25 & 85.00 & +3.75 \\
 & 1 & 77.50 & 86.72 & +9.22 \\
 & 2 & 80.31 & 82.81 & +2.50 \\
\bottomrule
\end{tabular}
\end{table}

\paragraph{Learning progress and uncertainty.}
Per-seed endpoint values $E_{250}=S(250)$ are retained in
Table~\ref{tab:text-seeds} for completeness.
Normalized integrated success summarizes learning progress
over the complete raw evaluation grid $t_j=5j$:
\begin{equation}
I_s=\frac{1}{250}\sum_{j=1}^{50}
\frac{S_s(t_{j-1})+S_s(t_j)}{2}(t_j-t_{j-1}).
\label{eq:integrated-success}
\end{equation}
Table~\ref{tab:matched-readouts} reports success and integrated success with paired
intervals $\overline\Delta\pm4.30265\,s_\Delta/\sqrt3$.
The three training seeds determine uncertainty; evaluations along one
training trajectory are dependent. The $L_5$ and integrated-success
intervals span zero in all four Qwen2.5 settings.
These intervals quantify the uncertainty accompanying the observed
mean gains.

\begin{table}[!htbp]
\centering
\small
\setlength{\tabcolsep}{7pt}
\renewcommand{\arraystretch}{1.08}
\caption{Success and learning progress for GraphGPO and \method{}
across three training seeds.
Entries give mean\,$\pm$\,sample SD; both scores use a 0--100 scale.
$\Delta$ is \method{} minus
GraphGPO in percentage points. Success follows Eq.~\ref{eq:late5};
normalized raw-curve AUC summarizes learning progress.
Intervals are 95\% paired Student-$t$ intervals
($\mathrm{df}=2$); definitions appear in Appendix~\ref{app:result-freezing}.}
\label{tab:matched-readouts}
\begin{tabular}{@{}lrr@{}}
\toprule
Setting / readout & Success (\%) & Normalized AUC (\%) \\
\midrule
\multicolumn{3}{@{}l}{\textbf{ALFWorld / 1.5B}} \\
GraphGPO & $92.60\,\pm\,1.93$ & $77.42\,\pm\,1.73$ \\
\method{} & $93.65\,\pm\,1.40$ & $78.68\,\pm\,1.34$ \\
Paired $\Delta$ (pp) & $+1.04$ & $+1.25$ \\
95\% paired interval & $[-0.87,\,+2.96]$ & $[-0.67,\,+3.18]$ \\
\midrule
\multicolumn{3}{@{}l}{\textbf{ALFWorld / 7B}} \\
GraphGPO & $94.84\,\pm\,2.11$ & $89.02\,\pm\,1.42$ \\
\method{} & $96.30\,\pm\,0.86$ & $90.48\,\pm\,0.87$ \\
Paired $\Delta$ (pp) & $+1.46$ & $+1.46$ \\
95\% paired interval & $[-5.59,\,+8.50]$ & $[-0.21,\,+3.13]$ \\
\midrule
\multicolumn{3}{@{}l}{\textbf{WebShop / 1.5B}} \\
GraphGPO & $77.03\,\pm\,1.22$ & $66.00\,\pm\,0.85$ \\
\method{} & $78.13\,\pm\,2.97$ & $68.98\,\pm\,1.62$ \\
Paired $\Delta$ (pp) & $+1.09$ & $+2.98$ \\
95\% paired interval & $[-4.04,\,+6.23]$ & $[-3.04,\,+8.99]$ \\
\midrule
\multicolumn{3}{@{}l}{\textbf{WebShop / 7B}} \\
GraphGPO & $80.73\,\pm\,1.52$ & $71.21\,\pm\,1.15$ \\
\method{} & $81.20\,\pm\,5.02$ & $71.18\,\pm\,1.12$ \\
Paired $\Delta$ (pp) & $+0.47$ & $-0.02$ \\
95\% paired interval & $[-15.30,\,+16.24]$ & $[-4.68,\,+4.64]$ \\
\bottomrule
\end{tabular}
\end{table}

\subsection{Matched training-transition budgets at 1.5B}
\label{app:transition-budget-1p5b}

This retrospective diagnostic covers the twelve Qwen2.5-1.5B runs:
\method{} and GraphGPO on ALFWorld and WebShop, each with training seeds
0, 1, and 2. The effective configurations disable dynamic rollout
filtering and collect $16\times8=128$ trajectories per update.
The logged mean episode length $\overline L_{m,s,u}$ counts active
environment steps per unique trajectory. Actor padding duplicates rows,
while this metric deduplicates them by trajectory identity. We reconstruct
cumulative training transitions as
\begin{equation}
B_{m,s}(t)=\sum_{u=1}^{t}
\operatorname{round}\!\left(128\,\overline L_{m,s,u}\right).
\label{eq:training-transition-budget}
\end{equation}
With 128 trajectories, the three-decimal logged mean uniquely determines
the integer total; every reconstructed update passes the rounding check.
We follow the retained checkpoint lineage across resumes. Validation,
reload-only evaluation, rolled-back updates, and unfinished failed
attempts are outside this training budget.

For each environment and seed, we enumerate the scheduled checkpoint
pairs $t_M,t_G\in\{5,10,\ldots,250\}$ satisfying
\begin{equation}
\frac{|B_{M,s}(t_M)-B_{G,s}(t_G)|}
{\min\{B_{M,s}(t_M),B_{G,s}(t_G)\}}<0.01,
\label{eq:transition-budget-match}
\end{equation}
where $M$ and $G$ denote \method{} and GraphGPO. We maximize the smaller
of the two budgets, breaking ties by smaller relative mismatch and then
larger combined update count. Selection uses the interaction counts alone.
The plotted readout is raw validation success at the selected checkpoint,
aggregated equally over the three training seeds. The budget labels average
the paired costs across seeds. These checkpoint readouts are distinct
from the primary $L_5$ metric.

\begin{figure}[!htbp]
\centering
\includegraphics[width=0.90\linewidth]{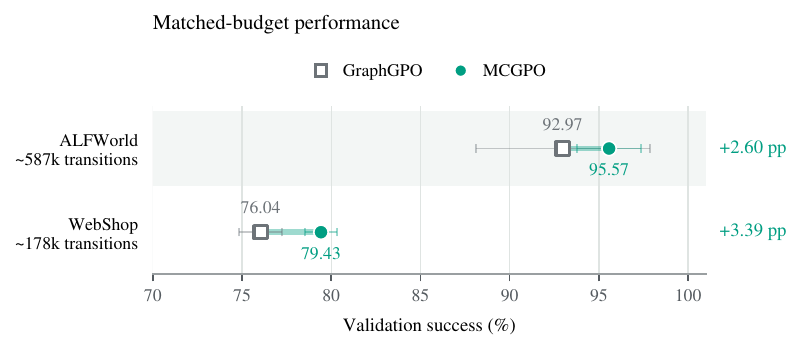}
\caption{Qwen2.5-1.5B validation success at nearly matched training-transition
budgets. Markers show three-seed means and whiskers sample SD. Each seed's
checkpoint pair differs in budget by less than 1\%; displayed budgets
average the paired costs. Counts cover retained training interactions;
Appendix~\ref{app:transition-budget-1p5b} specifies matching and restart accounting.}
\label{fig:matched-transition-budget-1p5b}
\end{figure}

\begin{table}[!htbp]
\centering
\small
\setlength{\tabcolsep}{3pt}
\renewcommand{\arraystretch}{1.12}
\caption{Selected budget-matched checkpoints for Qwen2.5-1.5B.
Slash-separated entries give GraphGPO/\method{}. Gap is the relative
budget difference in Eq.~\ref{eq:transition-budget-match}; $\Delta$ is
\method{} minus GraphGPO success in percentage points.}
\label{tab:transition-budget-1p5b}
\begin{tabular}{@{}lcrrrrr@{}}
\toprule
Env. & Seed & Updates & Transitions & Gap (\%) & Success (\%) & $\Delta$ (pp) \\
\midrule
ALFWorld & 0 & 230/250 & 594,465/592,472 & 0.336 & 87.50/97.66 & +10.16 \\
 & 1 & 250/250 & 575,983/581,151 & 0.897 & 96.88/94.53 & -2.34 \\
 & 2 & 245/250 & 590,739/585,967 & 0.814 & 94.53/94.53 & 0.00 \\
\midrule
WebShop & 0 & 250/215 & 174,190/173,006 & 0.684 & 75.00/80.47 & +5.47 \\
 & 1 & 250/230 & 188,293/187,096 & 0.640 & 77.34/78.91 & +1.56 \\
 & 2 & 250/200 & 172,791/171,773 & 0.593 & 75.78/78.91 & +3.13 \\
\bottomrule
\end{tabular}
\end{table}

Mean gains are 2.60 points on ALFWorld and 3.39 on WebShop. WebShop has
three positive paired differences; ALFWorld has one positive, one zero,
and one negative difference. Descriptive paired 95\% Student-$t$ intervals
($\mathrm{df}=2$) are $[-13.90,19.11]$ and $[-1.50,8.27]$ points for
ALFWorld and WebShop, respectively.

\subsection{Mechanism controls}
\label{app:mechanism-readouts}
Table~\ref{tab:mechanisms}(a) reports one completed Qwen2.5-1.5B run per
variant and environment. The candidate-progress control uses candidate
trajectories' progress in place of opposite-fold terminal outcomes,
jointly changing evidence source, labels, support, and weights.
The loss-removal control sets $\lambda_{\rm pair}=0$ while retaining
construction diagnostics. These single-run comparisons evaluate the
specified interventions; replication would estimate their variability.

The rubric-shaping control keeps GraphGPO credit and adds
\begin{equation}
\Delta r_t=0.05\,N_t^{\rm increase}-0.05\,N_t^{\rm decrease}
\label{eq:rubric-event-shaping}
\end{equation}
to the executed response's last valid token before advantage computation,
with pairwise learning disabled. Counts include changes in graded and
count-valued channels; a channel can increase more than once.
This Qwen2.5 comparison evaluates the event-shaping rule at a fixed
coefficient.

\paragraph{Factorial controls of evidence source and weighting.}
\label{app:factorial-controls}
Table~\ref{tab:mechanisms}(b) crosses two factors on Qwen3-4B/WebShop:
opposite-fold versus same-fold terminal evidence, and evidence-based versus
uniform loss weights. All four variants share initialization, training data,
action-only prompts with thinking disabled, and 250 updates with 16 task
groups of eight trajectories per update. We use seeds 0, 1, and 2, learning
rate $10^{-6}$, and fixed folds $\{0,2,4,6\}$ and $\{1,3,5,7\}$.
The construction parameters in Table~\ref{tab:preference-settings} and
$\lambda_{\rm pair}=0.05$ are shared, without variant-specific tuning.

A (Full) and C use the four opposite-fold trajectories as terminal evidence;
B and D use the four same-fold trajectories, including the candidates
themselves. A and B retain the original evidence-weight formula.
C and D assign equal loss weights to the selected preferences while
preserving their original total weight. Matching rules, support thresholds,
and pair-selection criteria otherwise remain fixed.

Table~\ref{tab:qwen3-factorial} reports individual seeds and paired contrasts.
Opposite-fold evidence improves success under both evidence weighting
($A-B$: +\factorFoldWeightedGain{} pp) and uniform weighting
($C-D$: +\factorFoldUniformGain{} pp). Evidence weighting improves success
with both opposite-fold ($A-C$: +\factorWeightOppositeGain{} pp) and same-fold
evidence ($B-D$: +\factorWeightSameGain{} pp). All four contrasts are positive
in every seed, and their paired 95\% intervals lie above zero. These results
support each design choice under both settings of the other factor.

\begin{table}[!htbp]
\centering\small
\setlength{\tabcolsep}{5pt}
\renewcommand{\arraystretch}{1.1}
\caption{Qwen3-4B WebShop factorial controls from Table~\ref{tab:mechanisms}(b).
(a) Per-seed success (\%), averaging the final five scheduled evaluations.
(b) Paired gains in percentage points, with 95\% Student-$t$ intervals
over three seed differences ($\mathrm{df}=2$). Statistics precede display rounding.}
\label{tab:qwen3-factorial}
\begin{tabular*}{0.86\linewidth}{@{\extracolsep{\fill}}lrrr@{}}
\multicolumn{4}{@{}l}{\textbf{(a) Per-seed task success (\%)}} \\[2pt]
\toprule
Configuration & Seed 0 & Seed 1 & Seed 2 \\
\midrule
A: \method{} (Full) & 85.63 & 78.59 & 82.19 \\
B: Same-fold terminal & 83.28 & 76.56 & 79.53 \\
C: Uniform weighting & 84.22 & 76.72 & 80.63 \\
D: Same-fold + uniform & 81.56 & 74.84 & 77.97 \\
\bottomrule
\end{tabular*}
\par\vspace{0.6em}
\begin{tabular*}{0.86\linewidth}{@{\extracolsep{\fill}}lrr@{}}
\multicolumn{3}{@{}l}{\textbf{(b) Paired component gains}} \\[2pt]
\toprule
Contrast & Mean gain (pp) & 95\% paired CI \\
\midrule
$A-B$ & +2.34 & $[+1.57, +3.12]$ \\
$A-C$ & +1.61 & $[+1.02, +2.21]$ \\
$C-D$ & +2.40 & $[+1.28, +3.52]$ \\
$B-D$ & +1.67 & $[+1.44, +1.89]$ \\
\bottomrule
\end{tabular*}
\end{table}

\paragraph{Scorer-matched Qwen3 comparison.}
\label{app:scorer-matched-control}
This experiment compares two ways of using the same scorer-derived
supervision, testing whether the complete \method{} preference-learning
scheme improves on reward shaping.
For Qwen3-4B/8B, GraphGPO + shaping uses all scorer information available to
\method{} (Appendix~\ref{app:matching-coordinates}) through training rewards.
Table~\ref{tab:qwen3-shaping-paired} reports the three-method comparison
and paired gains across seeds 0, 1, and 2. \method{} improves mean success
over shaping by \qwenFourShapingGain{} points at 4B and
\qwenEightShapingGain{} at 8B, with positive gains in all six paired runs.
These results support the complete \method{} scheme as a more effective
use of the scorer signals than the compared reward-shaping alternative.

For differences $d_s$, the paired 95\%
Student-$t$ interval is $\bar d \pm t_{0.975,2}\,s_d/\sqrt{3}$, where
$s_d$ is their sample SD. It summarizes training-seed variability under
the paired-$t$ approximation; both intervals lie above zero.

\begin{table}[!htbp]
\centering\small
\setlength{\tabcolsep}{5pt}
\renewcommand{\arraystretch}{1.1}
\caption{WebShop comparison with a scorer-matched shaping control.
(a) Task success (\%), reported as mean $\pm$ sample SD over three seeds.
(b) \method{} gains over GraphGPO + shaping, in percentage points.
Paired 95\% Student-$t$ intervals
($\mathrm{df}=2$) use the per-seed differences shown.}
\label{tab:qwen3-shaping-paired}
\begin{tabular*}{0.86\linewidth}{@{\extracolsep{\fill}}lrr@{}}
\multicolumn{3}{@{}l}{\textbf{(a) Task success (\%)}} \\[2pt]
\toprule
Method & Qwen3-4B & Qwen3-8B \\
\midrule
GraphGPO & $78.18 \pm 2.62$ & $79.69 \pm 1.95$ \\
GraphGPO + shaping & $79.85 \pm 3.21$ & $82.19 \pm 1.44$ \\
\method{} & $\mathbf{82.14} \pm 3.52$ & $\mathbf{84.84} \pm 1.96$ \\
\bottomrule
\end{tabular*}
\par\vspace{0.6em}
\begin{tabular*}{0.86\linewidth}{@{\extracolsep{\fill}}lrrrrr@{}}
\multicolumn{6}{@{}l}{\textbf{(b) Paired gains over GraphGPO + shaping (pp)}} \\[2pt]
\toprule
Backbone & Seed 0 & Seed 1 & Seed 2 & Mean gain & 95\% paired CI \\
\midrule
Qwen3-4B & +2.50 & +1.87 & +2.50 & +2.29 & $[+1.39, +3.19]$ \\
Qwen3-8B & +2.50 & +3.28 & +2.19 & +2.66 & $[+1.26, +4.05]$ \\
\bottomrule
\end{tabular*}
\end{table}

\subsection{Representation controls}
\label{app:representation-controls}
State-only, Rubric-only, and Full use the state coordinate, rubric
coordinates, and their joint context, respectively, with the same
preference construction and actor objective. Table~\ref{tab:representation-250}
reports seeds 0, 1, and 2 in all four settings. Full has the highest mean
under both readouts throughout, exceeding State-only by
\repStateMin{}--\repStateMax{} and Rubric-only by
\repRubricMin{}--\repRubricMax{} percentage points on the primary readout.
Full and State-only produce different selected preference sets during
training. Joint keys retain revisits after progress changes and condition
candidate actions and opposite-fold evidence on both state and progress.
State-only can merge different progress stages; Rubric-only can merge
different environmental situations. The gains support joint context for
selecting the auxiliary learning signal. WebShop State-only already
includes task-match score bins, so this control compares how structured
training information is used. Appendix~\ref{app:resolution-census}
illustrates visit retention and support in a separate diagnostic sample.

\begin{table}[!htbp]
\centering
\small
\setlength{\tabcolsep}{5pt}
\renewcommand{\arraystretch}{1.08}
\caption{Supplementary representation controls at 250 updates.
Full, State-only, and Rubric-only use seeds 0, 1, and 2 in every
setting. Entries give mean success (\%) and sample SD across the
three seeds. $L_5$ averages the final five
scheduled evaluations, and $E_{250}$ is the terminal evaluation.
State-only retains the environment-specific fields of its state
abstraction, including WebShop's match-score bins.}
\label{tab:representation-250}
\begin{tabular}{@{}llcrr@{}}
\toprule
Setting & Comparison key & $n$ & $L_5$ & $E_{250}$ \\
\midrule
ALFWorld / 1.5B & \method{} & 3 & $93.65\,\pm\,1.40$ & $95.57\,\pm\,1.80$ \\
 & State-only & 3 & $85.57\,\pm\,2.32$ & $83.85\,\pm\,1.97$ \\
 & Rubric-only & 3 & $87.81\,\pm\,1.59$ & $87.76\,\pm\,1.80$ \\
\midrule
ALFWorld / 7B & \method{} & 3 & $96.30\,\pm\,0.86$ & $96.09\,\pm\,1.35$ \\
 & State-only & 3 & $92.34\,\pm\,1.25$ & $91.93\,\pm\,1.97$ \\
 & Rubric-only & 3 & $91.15\,\pm\,1.83$ & $90.36\,\pm\,1.97$ \\
\midrule
WebShop / 1.5B & \method{} & 3 & $78.13\,\pm\,2.97$ & $75.26\,\pm\,7.00$ \\
 & State-only & 3 & $72.50\,\pm\,1.25$ & $73.18\,\pm\,1.19$ \\
 & Rubric-only & 3 & $70.42\,\pm\,0.94$ & $72.40\,\pm\,2.96$ \\
\midrule
WebShop / 7B & \method{} & 3 & $81.20\,\pm\,5.02$ & $80.21\,\pm\,7.09$ \\
 & State-only & 3 & $71.30\,\pm\,4.48$ & $69.27\,\pm\,7.26$ \\
 & Rubric-only & 3 & $72.19\,\pm\,2.99$ & $70.57\,\pm\,2.26$ \\
\bottomrule
\end{tabular}
\end{table}

\subsection{Learning curves and supervision dynamics}
\label{app:all-text-curves}

\begin{figure}[!htbp]
\centering
\includegraphics[width=\linewidth]{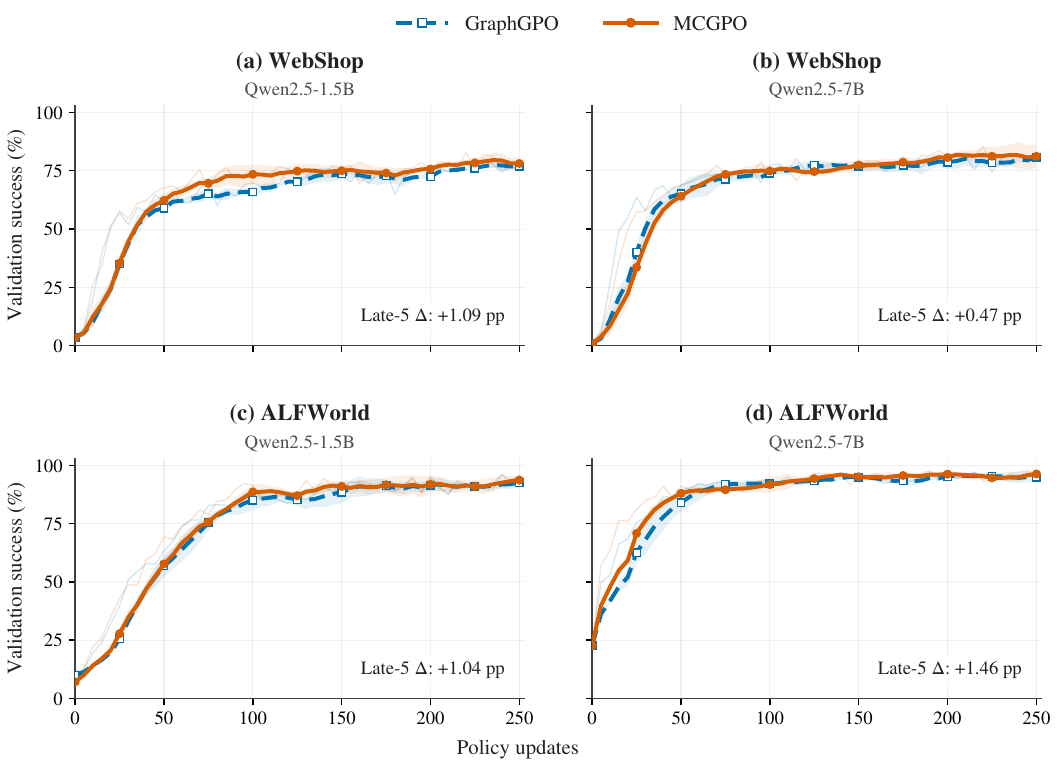}
\caption{Validation success in all four Qwen2.5 settings.
Each seed uses up to five trailing scheduled evaluations, including all
available points at the beginning. Lines and bands give the three-seed
mean and sample SD; faint lines retain raw means.
The fixed $L_5$ window is defined separately in Eq.~\ref{eq:late5}.}
\label{fig:all-text-curves}
\end{figure}

\begin{figure}[!htbp]
\centering
\includegraphics[width=\linewidth]{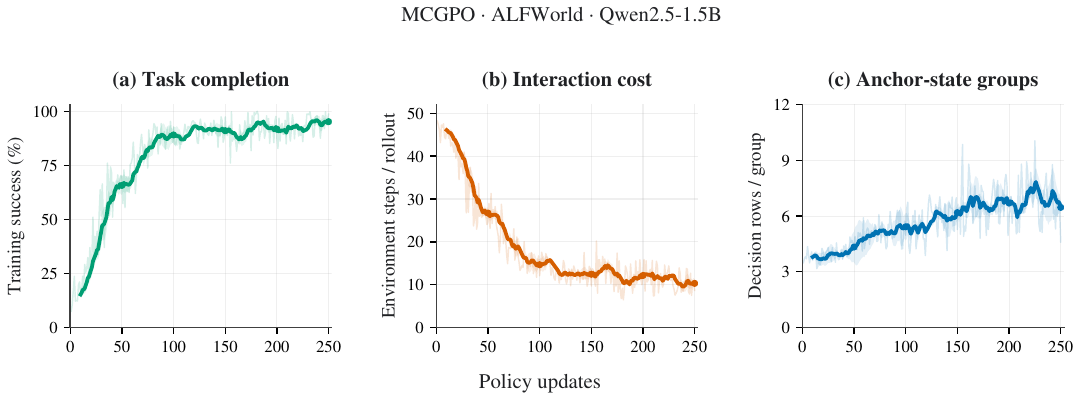}
\caption{ALFWorld Qwen2.5-1.5B training behavior:
(a) rollout success, (b) environment steps per rollout, and
(c) anchor-state group size in the task-credit branch.
Dark lines average ten trailing updates within each seed, then report
the three-seed mean; bands show sample SD and faint lines show seed means.}
\label{fig:rollout-dynamics}
\end{figure}

This subsection supplies the full curves referenced by RQ1 and the
aggregation definitions for RQ3. Figures~\ref{fig:all-text-curves}
and~\ref{fig:rollout-dynamics} describe policy learning and interaction
length over the same retained training runs.

\paragraph{Training behavior.}
The ALFWorld analysis includes all 750 updates from three completed Full
runs. Windows 26--50, 101--125, and 226--250 contain 25 updates per seed.
Mean training success rises from 60.3\% to 89.9\% and 95.3\%, while mean
rollout length falls from 29.34 to 13.91 and 10.17 steps.
Length counts active interactions per distinct trajectory, including
successful, failed, and truncated runs under the 50-step cap.
Mean anchor-state group size rises from 4.10 to 5.54 and 6.62; it counts
decision rows sharing task and observation, including repeated visits.
These are joint training trends; component effects are assessed in RQ2.

\paragraph{Preference supply and agreement.}
Figure~\ref{fig:preference-dynamics} uses all 750 WebShop 1.5B updates
in nonoverlapping 25-update windows. Each seed contributes equally to
the plotted means and sample SD. Counts include zero-pair updates;
outcome ratios pool $W,T,L$ within each seed/window before aggregation.
The pooled 74.35\% agreement instead weights all 1,990 accepted pairs.
Coarse matching supplies 1,805 pairs (90.70\%).

\paragraph{Weights and entry into learning.}
Evidence contrast and weight means first weight updates by accepted-pair
count within each seed/window. Their pooled means are 0.781 and 0.208.

Among all pairs, 1,547 have balanced evidence support one and 443 have
support two. The weight combines outcome contrast, support shrinkage,
and matching resolution. TensorBoard values matched to the retained
process and update confirm nonzero pairwise loss in all 511 updates
with accepted pairs; all 239 empty-pair updates have zero loss.
This establishes objective activation, complementing the loss-removal
control in RQ2. All these counts describe the constructor's selected
output. Source logs retain the exact count identities and checkpoint lineage.
\section{Rubric, matching, and preference diagnostics}
\label{app:resolution-census}

This section specifies the comparison keys used in RQ2 and supplies
diagnostics for RQ3. The rubric specification defines recorded task
progress; archived examples illustrate visit retention and matching
support, and WebShop outcome diagnostics characterize selected preferences.


\subsection{Environment rubric rules}
\label{app:rubric-rules}
Extractors run after each transition. The next decision uses that
transition's status as its pre-action $C$ and $F$, following the opening
convention in Appendix~\ref{app:matching-coordinates}.
ALFWorld reads task metadata and tracks target-object sets
$\mathcal A,\mathcal P,\mathcal L$ for acquisition, processing, and
placement. The required count is $r=2$ for Pick2 and $r=1$ otherwise.
Successful take, process, and placement feedback updates these sets;
entity matching preserves object-instance numbers.

\begin{table}[!htbp]
\centering\footnotesize
\setlength{\tabcolsep}{4pt}
\renewcommand{\arraystretch}{1.13}
\caption{ALFWorld channel definitions. Symbols identify the nodes in
Table~\ref{tab:rubric-dependency-edges}; $\mathrm{1}[\cdot]$ denotes an indicator.
The sets and required object count $r$ are defined in the text.}
\label{tab:rubric-channels-alfworld}
\begin{tabular}{@{}>{\raggedright\arraybackslash}p{0.30\linewidth}
>{\raggedright\arraybackslash}p{0.66\linewidth}@{}}
\toprule
Channel (symbol) & Status after the observed transition \\
\midrule
\texttt{target\_acquired} ($A$)
& $\min(r,|\mathcal A|)$; acquisition history persists after placement. \\
\texttt{target\_in\_hand} ($H$)
& Whether the worker's held object matches the target type. \\
\texttt{target\_processed} ($P$)
& $\mathrm{1}[\mathcal P\neq\emptyset]$; processing history persists after release. \\
\texttt{target\_placed} ($L$)
& $\min(r,|\mathcal L|)$; taking or relocating an object updates this set. \\
\texttt{light\_on} ($B$)
& Tracked target-light state: successful off commands/feedback set false;
on, use, or toggle set true unless off is indicated. \\
\texttt{goal\_progress} ($G$)
& Look: $H+B$. Clean/Heat/Cool:
$\mathrm{1}[\mathcal P\neq\emptyset]+\mathrm{1}[\mathcal L\neq\emptyset]
+\mathrm{1}[\mathcal P\cap\mathcal L\neq\emptyset]$.
Pick/Pick2: $\min(r,|\mathcal L|)$. \\
\texttt{solved} ($S$)
& The environment's \texttt{won} flag. \\
\bottomrule
\end{tabular}
\end{table}

WebShop reads browser-session fields and its existing goal scorer.
Product-match channels therefore use additional structured training
supervision, including scorer evaluation before purchase. Reusing
rollouts saves trajectory collection while retaining this information
requirement. The extractor's task contract omits the target product
identifier; the environment scorer retains its goal metadata.
The adapters use deterministic rules and the benchmark scorer.
With matched/requested attribute and option counts $M_a,M_o,N_a,N_o$,
type score $u$, and price predicate $p$, purchase readiness is
\begin{equation}
q_{\rm preview}=u\,\frac{M_a+M_o+p}{N_a+N_o+1}.
\label{eq:rubric-webshop-preview}
\end{equation}
Absent components and caught scoring errors retain initialized zeros.
This includes an omitted option component for a no-option goal.
Terminal evidence uses $y=\mathrm{1}[\mathrm{score}\ge0.999]$.

\begin{table}[!htbp]
\centering\footnotesize
\setlength{\tabcolsep}{4pt}
\renewcommand{\arraystretch}{1.13}
\caption{WebShop channel definitions. Symbols refer to WebShop nodes in
Table~\ref{tab:rubric-dependency-edges}. Match components come from the
environment goal scorer; default handling is specified in the text.}
\label{tab:rubric-channels-webshop}
\begin{tabular}{@{}>{\raggedright\arraybackslash}p{0.30\linewidth}
>{\raggedright\arraybackslash}p{0.66\linewidth}@{}}
\toprule
Channel (symbol) & Status after the observed transition \\
\midrule
\texttt{search\_issued} ($Q$)
& Whether the session has search keywords or its searched flag is true. \\
\texttt{product\_opened} ($O$)
& Whether a product is selected or the product-open flag is true. \\
\texttt{type\_match} ($T$)
& The scorer's type-match component \texttt{r\_type}. \\
\texttt{attribute\_match} ($A$)
& Requested-attribute match fraction; scorer \texttt{r\_att} is stored as
\texttt{r\_attr}. \\
\texttt{option\_match} ($U$)
& Requested-option match fraction \texttt{r\_option}; the standard worker
retains zero when the scorer omits this component. \\
\texttt{price\_ok} ($P$)
& The scorer's Boolean price predicate \texttt{r\_price}. \\
\texttt{purchase\_readiness} ($R$)
& Preview score of the selected product/options, computed before purchase
using Eq.~\ref{eq:rubric-webshop-preview}. \\
\texttt{purchased} ($B$)
& The environment's \texttt{done} flag. \\
\texttt{task\_score} ($K$)
& Raw environment score clipped to $[0,1]$, before binary reward remapping. \\
\texttt{solved} ($S$)
& True exactly when the environment terminates with raw score $1$. \\
\bottomrule
\end{tabular}
\end{table}

\begin{table}[!htbp]
\centering\footnotesize
\setlength{\tabcolsep}{4pt}
\renewcommand{\arraystretch}{1.13}
\caption{Complete dependency templates. Symbols use the respective
environment's channel table. E denotes \texttt{requires\_ever} and
C denotes \texttt{requires\_current}. Each ALFWorld family includes the
common edge; set notation expands to one edge per element.}
\label{tab:rubric-dependency-edges}
\begin{tabular}{@{}>{\raggedright\arraybackslash}p{0.23\linewidth}
>{\raggedright\arraybackslash}p{0.13\linewidth}
>{\raggedright\arraybackslash}p{0.58\linewidth}@{}}
\toprule
Environment/task & Relation & Directed edges \\
\midrule
ALFWorld: all & E & $A\to H$ \\
Pick / Pick2 & E & $A\to L$ \\
 & C & $L\to G$, $G\to S$ \\
Clean / Heat / Cool & E & $A\to P$, $A\to L$ \\
 & C & $P\to L$, $P\to G$, $L\to G$, $G\to S$ \\
Look & C & $H\to G$, $B\to G$, $H\to S$, $B\to S$, $G\to S$ \\
\midrule
WebShop: all & E & $Q\to O$ \\
 & C & $O\to\{T,A,U,P\}$ \\
 & C & $\{T,A,U,P\}\to R$ \\
 & C & $R\to B$, $B\to\{K,S\}$ \\
\bottomrule
\end{tabular}
\end{table}

Table~\ref{tab:rubric-dependency-edges} lists the task-rule graphs.
All non-blocking edges supply the prerequisite tests in
Eq.~\ref{eq:frontier-rules}; the text-environment graphs have no blocking
edges. ALFWorld's coarse budget bin partitions the remaining horizon
at $1/4,1/2,3/4$. WebShop's score bins use
$\operatorname{round}_{\rm even}(4\,\operatorname{clip}(v,0,1))$.
Action keys preserve canonical ALFWorld commands; WebShop uses
navigation/purchase markers and goal-aware search, product, and option
keys from the pre-action page, with goal-token overlap thresholds
$0.15,0.35,0.60$. Missing required trace fields disable
auxiliary supervision while retaining the task update.
Progress-measurement errors can change groups and action support;
extractor precision and recall against independent annotations remain
unmeasured.

\subsection{Matching support and a captured example}
\begin{figure}[!htbp]
\centering
\includegraphics[width=\linewidth]{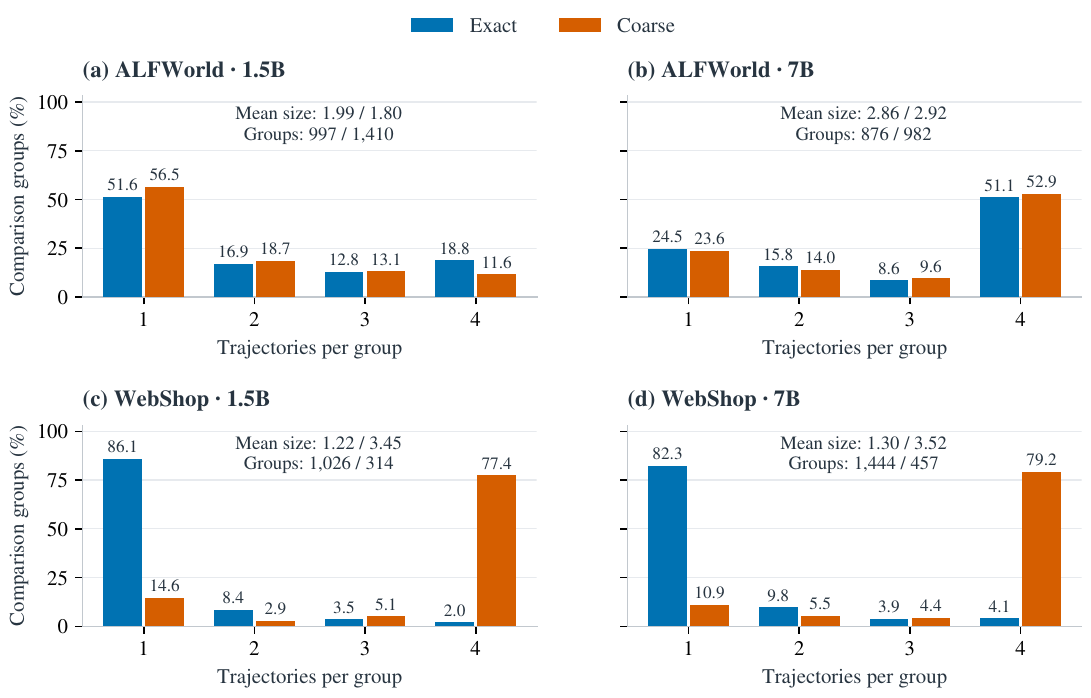}
\caption{First-visit group sizes across ten update-60 captures
(1.5B: seeds 1/2; 7B: seeds 0/1/2, in both environments).
Exact and coarse rules retain the same rubric coordinates; groups count
distinct contributors within a four-trajectory candidate fold.
Bars include singletons, and annotations report mean sizes and group counts.}
\label{fig:group-resolution}
\end{figure}

The census uses ten archived update-60 captures: 160 task instances
and 1,280 trajectories across the four text settings. These captures are
a separate diagnostic sample from the training runs in
Table~\ref{tab:representation-250}. Each candidate fold contains four
trajectories; groups count first eligible visits
per trajectory and key. Coverage divides records in groups of size at
least two by all retained records at that resolution.
Figure~\ref{fig:group-resolution} reports the distribution over groups;
Table~\ref{tab:resolution-support} reports coverage over records and counts
supported pairs. Both precede outcome filtering.

\begin{table}[!htbp]
\centering
\small
\setlength{\tabcolsep}{4pt}
\renewcommand{\arraystretch}{1.08}
\caption{Single-resolution comparison availability in all ten captured update-60 batches. Non-singleton coverage is the fraction of first-visit records in groups with at least two contributors. Supported pairs have distinct actions and at least one opposite-fold observation of each action. These counts precede outcome-contrast filtering and the final quota; exact and coarse counts overlap.}
\label{tab:resolution-support}
\begin{tabular}{@{}lrrrr@{}}
\toprule
& \multicolumn{2}{c}{Non-singleton records (\%)} & \multicolumn{2}{c}{Supported row pairs} \\
\cmidrule(lr){2-3}\cmidrule(lr){4-5}
Setting & Exact & Coarse & Exact & Coarse \\
\midrule
ALFWorld / 1.5B & 74.07 & 68.57 & 143 & 133 \\
ALFWorld / 7B & 91.43 & 91.90 & 86 & 252 \\
WebShop / 1.5B & 29.19 & 95.76 & 2 & 19 \\
WebShop / 7B & 36.61 & 96.89 & 14 & 79 \\
\bottomrule
\end{tabular}
\end{table}

\begin{figure}[!htbp]
\centering
\includegraphics[width=\linewidth]{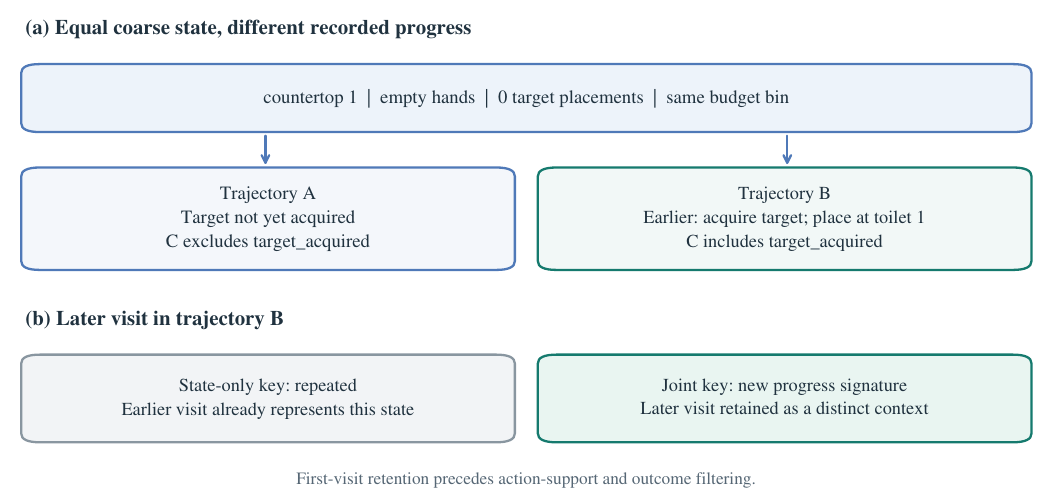}
\caption{Captured ALFWorld example: \emph{put two toiletpaper in drawer}.
Two trajectories reach countertop~1 with empty hands, zero target
placements, and the same budget bin. One previously acquired
toiletpaper~2 and placed it at toilet~1. Its changed progress signature
retains the later visit as a distinct context under the joint first-visit
rule, before preference selection.}
\label{fig:progress-retention}
\end{figure}

Exact and coarse are evaluated separately, so supported-pair counts can
overlap. On these archived snapshots, Full replay reproduces the stored
pairs and directions; State-only selects the same pairs, with no accepted
pairs in the WebShop subset after outcome filtering. These statistics
characterize local matching availability. Figure~\ref{fig:progress-retention}
illustrates how progress preserves a revisit as a distinct comparison
context; Table~\ref{tab:representation-250} assesses the benefit of applying
joint context throughout training.

\FloatBarrier
\subsection{WebShop agreement with candidate outcomes}
\label{app:transfer}
\begin{table}[!htbp]
\centering
\small
\setlength{\tabcolsep}{3.8pt}
\renewcommand{\arraystretch}{1.08}
\caption{Archived WebShop training-trace diagnostics for Full, pooled over three seeds per Qwen2.5 backbone. $W/T/L$ denotes counts of preferred-candidate wins, ties, and reversed outcomes; agreement is $(W+T/2)/(W+T+L)$. Active gives accepted-pair updates / audited updates; Exact is the fraction of accepted pairs using exact matching. All counts refer to the retained trace sample.}
\label{tab:verifier}
\begin{tabular}{@{}lrrrrr@{}}
\toprule
Setting & Pairs & $W/T/L$ & Agreement (\%) & Active & Exact (\%) \\
\midrule
WebShop / 1.5B & 1428 & 771/568/89 & 73.88 & 330/450 & 8.3 \\
WebShop / 7B & 1425 & 633/693/99 & 68.74 & 319/450 & 20.1 \\
\bottomrule
\end{tabular}
\end{table}

Table~\ref{tab:verifier} pools WebShop $W/T/L$ counts of 1404/1261/188.
Agreement measures association with
candidate terminal outcomes after preference construction; RQ2 supplies
the policy controls. These archived samples contain repeated training
observations; the complete 250-update WebShop 1.5B diagnostics and
objective-activation check appear in Appendix~\ref{app:all-text-curves}.

\section{Limitations}
\label{app:limitations}

\method{} is evaluated on ALFWorld and WebShop with Qwen backbones.
Transfer to other domains requires suitable state summaries, rubric
channels, and task rules; WebShop also uses structured goal-scorer
information during training. Preferences are inferred from finite groups
of observational rollouts: coarse matching can retain outcome-relevant
differences, and terminal success depends on subsequent decisions as well
as the matched action. Evidence weighting moderates the influence of
sparse comparisons, while direct validation of local action preferences
remains an important direction. Our main comparisons use three training
seeds and repeated evaluation on a fixed validation set, so small gains
retain statistical uncertainty. Evaluation with broader model families,
more seeds, and independent test tasks would further assess generalization
and statistical precision.

\section{Environment Prompts}
\label{app:prompts}

GraphGPO and \method{} use the same environment templates within each
backbone setting. Each filled template forms a user message in the
backbone's chat template; rubric coordinates are computed by the training
adapters. The boxes below give the interaction prompts, including their
response instructions. Only line wrapping, trailing whitespace, and the
typographic hyphen in \emph{e-commerce} are normalized.

Placeholders in braces are filled from the current environment state.
The first decision omits the history block; later decisions include up to
the two most recent observation--action pairs, as specified in
Table~\ref{tab:setup}. Each history entry has the form
\texttt{[Observation n: 'observation', Action n: 'action']}.
\texttt{step\_count} counts completed actions and
\texttt{current\_step=step\_count+1}. A recorded ALFWorld example appears
in Appendix~\ref{app:trace}.

\subsection{ALFWorld}

ALFWorld provides the current text observation and admissible-action set.
The task appears in the reset observation and is inserted explicitly as
\verb|{task_description}| at subsequent decisions. The admissible-action
list excludes the help command. Both Qwen2.5 backbones use the following
reasoning-and-action prompts.

\begin{promptbox}{ALFWorld Interaction Prompt: first decision}
You are an expert agent operating in the ALFRED Embodied Environment.
Your current observation is: {current_observation}
Your admissible actions of the current situation are: [{admissible_actions}].

Now it's your turn to take an action.
You should first reason step-by-step about the current situation. This reasoning process MUST be enclosed within <think> </think> tags.
Once you've finished your reasoning, you should choose an admissible action for current step and present it within <action> </action> tags.
\end{promptbox}

\begin{promptbox}{ALFWorld Interaction Prompt: subsequent decisions}
You are an expert agent operating in the ALFRED Embodied Environment. Your task is to: {task_description}
Prior to this step, you have already taken {step_count} step(s). Below are the most recent {history_length} observations and the corresponding actions you took: {action_history}
You are now at step {current_step} and your current observation is: {current_observation}
Your admissible actions of the current situation are: [{admissible_actions}].

Now it's your turn to take an action.
You should first reason step-by-step about the current situation. This reasoning process MUST be enclosed within <think> </think> tags.
Once you've finished your reasoning, you should choose an admissible action for current step and present it within <action> </action> tags.
\end{promptbox}

\subsection{WebShop}

The prompt includes the shopping goal, rendered page, and available
search/click actions. A search bar contributes
\texttt{search[<your query>]}; clickable elements contribute
\texttt{click[element]}. If a history-containing prompt exceeds
13,000 characters, the implementation falls back to the no-history
context while retaining the current goal and observation. The following
complete prompts apply to Qwen2.5-1.5B/7B.

\begin{promptbox}{WebShop Interaction Prompt: first decision (Qwen2.5)}
You are an expert autonomous agent operating in the WebShop e-commerce environment.
Your task is to: {task_description}.
Your current observation is: {current_observation}.
Your admissible actions of the current situation are:
[
{available_actions}
].

Now it's your turn to take one action for the current step.
You should first reason step-by-step about the current situation, then think carefully which admissible action best advances the shopping goal. This reasoning process MUST be enclosed within <think> </think> tags.
Once you've finished your reasoning, you should choose an admissible action for current step and present it within <action> </action> tags.
\end{promptbox}

\begin{promptbox}{WebShop Interaction Prompt: subsequent decisions (Qwen2.5)}
You are an expert autonomous agent operating in the WebShop e-commerce environment.
Your task is to: {task_description}.
Prior to this step, you have already taken {step_count} step(s). Below are the most recent {history_length} observations and the corresponding actions you took: {action_history}
You are now at step {current_step} and your current observation is: {current_observation}.
Your admissible actions of the current situation are:
[
{available_actions}
].

Now it's your turn to take one action for the current step.
You should first reason step-by-step about the current situation, then think carefully which admissible action best advances the shopping goal. This reasoning process MUST be enclosed within <think> </think> tags.
Once you've finished your reasoning, you should choose an admissible action for current step and present it within <action> </action> tags.
\end{promptbox}

Qwen3-4B/8B uses the same task, observation, admissible-action, and history
fields with an action-only response instruction. The complete first-decision
and subsequent-decision templates follow. Both use the backbone's chat
template with \texttt{enable\_thinking=False}.

\begin{promptbox}{WebShop Interaction Prompt: first decision (Qwen3)}
You are an expert autonomous agent operating in the WebShop e-commerce environment.
Your task is to: {task_description}.
Your current observation is: {current_observation}.
Your admissible actions of the current situation are:
[
{available_actions}
].

Now it's your turn to take one action for the current step.
Choose exactly one action from the admissible-action list. Respond with exactly one complete <action>selected action</action> block and no other text.
\end{promptbox}

\begin{promptbox}{WebShop Interaction Prompt: subsequent decisions (Qwen3)}
You are an expert autonomous agent operating in the WebShop e-commerce environment.
Your task is to: {task_description}.
Prior to this step, you have already taken {step_count} step(s). Below are the most recent {history_length} observations and the corresponding actions you took: {action_history}
You are now at step {current_step} and your current observation is: {current_observation}.
Your admissible actions of the current situation are:
[
{available_actions}
].

Now it's your turn to take one action for the current step.
Choose exactly one action from the admissible-action list. Respond with exactly one complete <action>selected action</action> block and no other text.
\end{promptbox}
\section{Illustrative Task and Rollout Trace}
\label{app:trace}

This complete ALFWorld rollout uses \method{} with Qwen2.5-1.5B-Instruct
(seed 2, update 60, sibling slot 4, zero-based). The task requires placing
two distinct objects into a drawer. We select the shortest successful
sibling of this previously audited task for compactness. It illustrates
the interaction protocol, not a representative performance estimate or
a matched preference pair.

\begin{casebox}{ALFWorld task: recorded reset observation}
-= Welcome to TextWorld, ALFRED! =-

You are in the middle of a room. Looking quickly around you, you see a countertop 1, a drawer 8, a drawer 7, a drawer 6, a drawer 5, a drawer 4, a drawer 3, a drawer 2, a drawer 1, a garbagecan 1, a handtowelholder 1, a sinkbasin 2, a sinkbasin 1, a toilet 1, a toiletpaperhanger 1, and a towelholder 1.

Your task is to: put two toiletpaper in drawer.
\end{casebox}

This reset observation fills the initial template in
Appendix~\ref{app:prompts}; later decisions use the latest observation,
admissible actions, and up to two preceding observation--action pairs.
The box preserves all 12 parsed environment commands, including the
repeated pickup and replacement at steps 6--7 and the sink visit at step 8.
Feedback is verbatim, omitting appended location, inventory, move-history,
and action-list annotations. Nonterminal feedback enters the next observation;
\textrm{success} is the literal terminal feedback.

\begin{rolloutbox}{Complete rollout: put two toiletpaper in drawer}
\textbf{Step 1.} Action: go to countertop 1\\*
Feedback: You arrive at countertop 1. On the countertop 1, you see a cloth 1, a soapbottle 1, a spraybottle 1, a toiletpaper 3, and a toiletpaper 2.

\textbf{Step 2.} Action: take toiletpaper 2 from countertop 1\\*
Feedback: You pick up the toiletpaper 2 from the countertop 1.

\textbf{Step 3.} Action: go to drawer 1\\*
Feedback: You arrive at drawer 1. The drawer 1 is closed.

\textbf{Step 4.} Action: open drawer 1\\*
Feedback: You open the drawer 1. The drawer 1 is open. In it, you see nothing.

\textbf{Step 5.} Action: move toiletpaper 2 to drawer 1\\*
Feedback: You move the toiletpaper 2 to the drawer 1.

\textbf{Step 6.} Action: take toiletpaper 2 from drawer 1\\*
Feedback: You pick up the toiletpaper 2 from the drawer 1.

\textbf{Step 7.} Action: move toiletpaper 2 to drawer 1\\*
Feedback: You move the toiletpaper 2 to the drawer 1.

\textbf{Step 8.} Action: go to sinkbasin 1\\*
Feedback: You arrive at sinkbasin 1. On the sinkbasin 1, you see a sink 1, and a soapbar 2.

\textbf{Step 9.} Action: go to toilet 1\\*
Feedback: You arrive at toilet 1. On the toilet 1, you see a candle 1, a soapbar 3, and a toiletpaper 4.

\textbf{Step 10.} Action: take toiletpaper 4 from toilet 1\\*
Feedback: You pick up the toiletpaper 4 from the toilet 1.

\textbf{Step 11.} Action: go to drawer 1\\*
Feedback: You arrive at drawer 1. The drawer 1 is open. In it, you see a toiletpaper 2.

\textbf{Step 12.} Action: move toiletpaper 4 to drawer 1\\*
Feedback: success
\end{rolloutbox}

\paragraph{Feedback and outcome.}
Steps 1--11 receive reward 0; step 12 receives reward 10 and terminal
success label $Y=1$. Post-action placement counts are 1, 0, 1, and 2
at steps 5, 6, 7, and 12, respectively. The acquisition count stays at one
while \textrm{toiletpaper 2} is handled repeatedly, then reaches two when
\textrm{toiletpaper 4} is acquired at step 10. Matching uses progress
available before the current action.

\end{document}